%% file: main.tex
\documentclass[11pt]{article}

\usepackage[acronym]{glossaries}
\makeglossaries
\input{glossary/glossary.tex}

\usepackage[margin=1in]{geometry}
\usepackage{txfonts}
\usepackage[T1]{fontenc}
\usepackage[dvipsnames]{xcolor}
\usepackage[mathscr]{eucal}

\usepackage{enumitem} 
\usepackage{booktabs} 
\usepackage{array} 

\usepackage{parskip}
\setlist[itemize]{noitemsep, nolistsep}
\setlist[enumerate]{noitemsep, nolistsep}

\usepackage{tabularx}
\usepackage{colortbl}
\usepackage{diagbox}

\usepackage{tikz}
\usepackage{sankey}

\usepackage[toc]{appendix}

\usetikzlibrary{calc}

\usepackage{makecell} 

\usetikzlibrary{arrows,shapes,snakes,automata,backgrounds,petri,positioning} 

\newcommand{\colorA}{blue}
\newcommand{\colorB}{red}

\title{Evaluating the Energy Consumption of Machine Learning: Systematic Literature Review and Experiments}

\usepackage{authblk}
\author[1, 2]{Charlotte Rodriguez}
\author[1]{Laura Degioanni}
\author[1]{Laetitia Kameni}
\author[1]{Richard Vidal}
\author[2]{Giovanni Neglia}

\affil[1]{Accenture Labs, Sophia Antipolis, France. Email: \{firstname.lastname\}@accenture.com}
\affil[2]{Inria Université Côte d'Azur, Sophia Antipolis, France. Email: \{firstname.lastname\}@inria.fr}

\date{\today}

\usepackage{xurl}
\usepackage{hyperref}
\hypersetup{
    hidelinks,
	colorlinks,
	citecolor=blue!85!black,
	filecolor=black,
	linkcolor=blue!85!black,
	urlcolor=green!60!black
}

\usepackage[numbers]{natbib}
\usepackage{graphicx}

\usepackage{longtable}

\begin{document}
\maketitle

\begin{abstract}
    Monitoring, understanding, and optimizing the energy consumption of Machine Learning (ML) are various reasons why it is necessary to evaluate the energy usage of ML. However, there exists no universal tool that can answer this question for all use cases, and there may even be disagreement on how to evaluate energy consumption for a specific use case. Tools and methods are based on different approaches, each with their own advantages and drawbacks, and they need to be mapped out and explained in order to select the most suitable one for a given situation.
    We address this challenge through two approaches. First, we conduct a systematic literature review of all tools and methods that permit to evaluate the energy consumption of ML (both at training and at inference), irrespective of whether they were originally designed for machine learning or general software. Second, we develop and use an experimental protocol to compare a selection of these tools and methods. The comparison is both qualitative and quantitative on a range of ML tasks of different nature (vision, language) and computational complexity.
    The systematic literature review serves as a comprehensive guide for understanding the array of tools and methods used in evaluating energy consumption of ML, for various use cases going from basic energy monitoring to consumption optimization. Two open-source repositories are provided for further exploration. The first one contains tools that can be used to replicate this work or extend the current review. The second repository houses the experimental protocol, allowing users to augment the protocol with new ML computing tasks and additional energy evaluation tools.
\end{abstract}

\clearpage
\tableofcontents
\clearpage



\section{Introduction}

Reducing the energy consumption of Machine Learning (ML) and Software in general has many motivations. Besides the environmental impact of computing, other factors include the actual cost of energy \cite{corda2022}, and the energy limitations of battery powered systems such as embedded or mobile devices \cite{dariol2023, rieger2017, gauen2017}. Most of the studies reviewed here express concern about the current and future growth of the \acrfull{ict} energy consumption and carbon footprint, with data centers being the fastest growing source of emissions in the \acrshort{ict} sector \cite{fu2018}.
%
Regarding Artificial Intelligence (AI) and ML, the study \cite{strubell2019} examined the environmental impact of \acrfull{nlp} tasks, such as Bert, Transformer, and GPT-2 training, or \acrfull{nas}, and attracted considerable attention in 2019. The authors find that training BERT-base (one of the the smaller versions of BERT) produced about 652 kg of carbon dioxide equivalents (CO2eq), which is equivalent to the emissions of a round-trip flight between New York and San Francisco per passenger \cite{strubell2019, budennyy2022}. They also estimate that the neural architecture search and training of the Transformer T2T has an impact equivalent to five times that of a car life time (including fuel).

The increasing attention to the energy impact of \acrshort{ict} let researchers addressing the energy efficiency of \acrshort{ict} to shift the focus from maximizing performance based on physical capabilities to minimizing energy and carbon costs while maintaining the same level of performance, giving rise to the concept of ``Green \acrshort{ict}'' \cite{hankel2016}. Similar considerations have also emerged in AI, with the notion of ``Green AI'' being introduced as an alternative to ``Red AI'' (although both are considered important). The latter aims to develop machine learning models with the highest accuracy, at the expense of massive computational power and energy consumption, whereas the former seeks to create models with lower computational power and fewer carbon emissions \cite{schwartz2020}.
Examples of approaches to improve energy efficiency of software may be found in \cite{mobius2014} and \cite[Appendix B]{shahid2021a}. The latter study notably stresses how essential accurate evaluation of energy consumption of an application execution is in order to minimize this consumption. In the field of machine learning, many studies ask that energy (and carbon) cost of ML is reported in addition to accuracy metrics, notably to increase awareness and incentivise energy efficient ML algorithms \cite{henderson2020}. Some machine conferences, such as ICML or NeurIPS, now ask contributions to declare the amount of compute and type of resources used (e.g., type of GPUs, type of platform) needed for their experiments. Studies also stress the need to render more available the reporting of energy and carbon metrics to the machine learning community \cite{garcia-martin2019a, garcia-martin2019b, henderson2020}, by easing the process of collecting these metrics and familiarising the community with the available approaches.
%



Our objective is to explore the different ways of evaluating the energy consumption of ML computing tasks, across all application domains. As we have seen above, tracking energy consumption is also a concern for computing tasks in general, this is why we also study ways of evaluating the energy consumption of software in general, thus also using terms such as ``software'' or ``application.''
Note that when assessing the total environmental impact of computing tasks, one should also take into account the impact of production and disposal of hardware (routers, computers, servers, for instance) \cite{bannour2021}. However, the latter impact is not in the scope of this work.
\subsection{Background}
\label{sec:background}

To evaluate the energy consumption (and carbon footprint) of a computing task, various methods and tools have been developed. Some are tailored for machine learning, while others can be used for general computing tasks. 
%
One can find several literature reviews and/or experimental comparison of such tools and methods. Among them the following four are particularly relevant for our interests: \cite{garcia-martin2019b, bannour2021, jay2023, fahad2019}.
Specifically, \cite{garcia-martin2019b} provides a comprehensive review of a broad set of energy consumption evaluation methods deemed applicable to machine learning, focusing on the methods based on building an estimation model of energy consumption by means of data observations (see Section \ref{sec:taxonomy} for detail). The two experimental studies \cite{bannour2021, jay2023} examine another set of evaluation methods this time based on basic estimation models for energy consumption and vendor specific interfaces to the CPU and GPU energy data (see Section \ref{sec:taxonomy} for detail), with the former study focusing on methods specifically developed for ML. Lastly, \cite{fahad2019} not only reviews but also experimentally tests all the approaches discussed in \cite{bannour2021, jay2023, garcia-martin2019b}. The authors found these methods to be insufficiently precise for their specific application, that is optimizing the dynamic energy consumption of software, which refers to the energy consumed solely due to the software’s execution \cite{fahad2019}. These studies are described in more detail in Section \ref{sec:presentation_secondary_studies}.
The existing literature draws several opposite conclusions. For instance, while \cite{bannour2021} concludes that the outputs of the tested tools vary significantly, \cite{jay2023} considers them to be relatively similar. 
Similarly, \cite{patterson2022} from Google comments the conclusions in \cite{strubell2019} about the environmental impact of large \acrshort{nlp} models, stating that ``Strubell et al.’s energy estimate for \acrshort{nas} ended up 18.7X too high for the
average organization (see \cite[Appendix C]{patterson2022}) and 88X off in emissions for energy-efficient organizations like Google.''
In summary, a good comprehension of what are the available approaches is still needed. Moreover, universally accepted energy evaluations are essential for informed decision-making in this domain.

\subsection{Research Question and Contributions}

The following research question is at the origin of this work: 
\begin{center}
\emph{What tools and methods currently permit to evaluate the energy consumption of machine learning computing tasks?}
\end{center}
Here, ``method'' entails that it is not yet implemented as a tool. Furthermore, the term ``evaluation'' includes both ``measuring'' and ``estimating,'' with or without the need to run the computing task. More precisely, for each discovered tool or method, we look to cover the following aspects of the research question: 
\begin{description}
    \item{[Approach]} What approach does this tool or method rely on?
    \item{[Context]} For what purpose, in what context has this tool or method been designed? 
    \item{[Constraints]} What are the constraints and limits of this tool or method?
\end{description}

Our contributions are the following ones.

\begin{figure}
    \centering
    \includegraphics[width = 10cm]{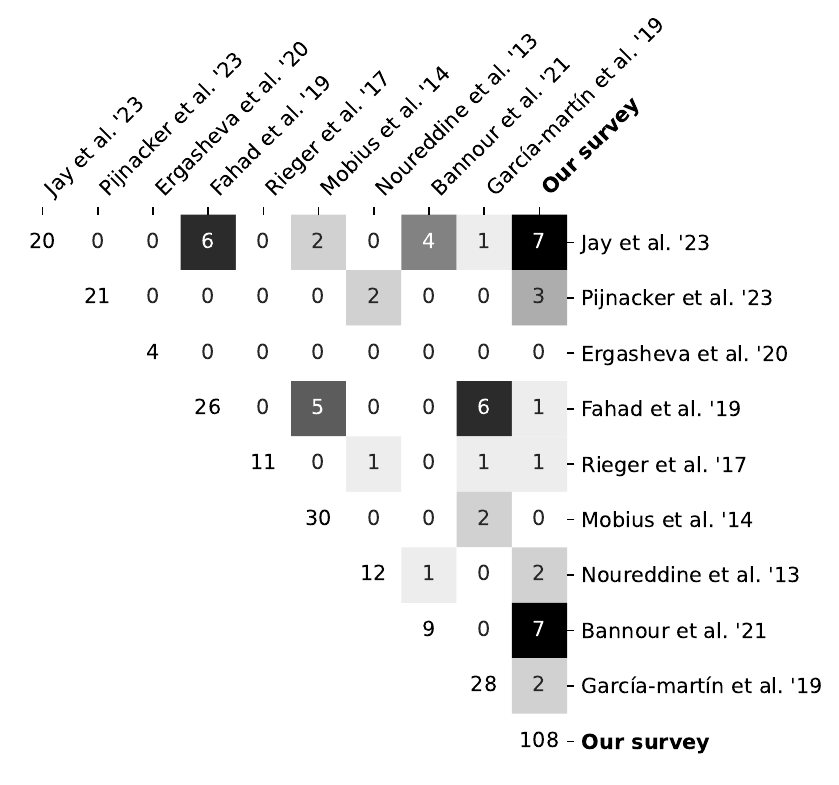}
    \caption{Intersection between our work and the surveys selected by our protocol:
Jay et al. '23 \cite{jay2023}, Pijnacker et al. '23 \cite{pijnacker2023}, Ergasheva et al. '20 \cite{ergasheva2020}, Fahad et al. '19 \cite{fahad2019}, Rieger et al. '17 \cite{rieger2017}, Mobius et al. '14 \cite{mobius2014}, Noureddine et al. '13 \cite{noureddine2013}, Bannour et al. '21 \cite{bannour2021}, García-martín et al. '19 \cite{garcia-martin2019a}.
    }
    \label{fig:heatmap}
\end{figure}

\emph{Contribution 1.} Our first contribution is that this work is the broadest review in terms of scope and number of studies reviewed, see Figure \ref{fig:heatmap}. This is due to the following three choices. Firstly, we propose a \acrfull{slr}, meaning that the search, selection, and analysis of studies are based on an a priori defined protocol, that we describe in detail in Section \ref{sec:protocol}. This is in contrast with the four works presented in Section \ref{sec:background} and additional surveys found during our review process, apart from \cite{bannour2021, pijnacker2023} that however have a narrower research scope. Secondly, the scope of our review includes all types of energy consumption evaluation approaches, and all application domains in the sense that we consider tools and methods developed not just for ML, but also for software in general (for monitoring, optimization, etc.). Thirdly, concerning ML application, we do not restrain ourselves to a specific subset of ML applications.

\emph{Contribution 2.} Our second contribution is an experimental comparison of evaluation tools and methods based on different approaches, on different ML computing tasks. 
For example, some tools and methods are based on direct measurements at the power outlet, others on vendor-specific sensors, and yet others on analytical estimation models (more detail on the different approaches will be provided in Section \ref{sec:taxonomy}), enabling us to observe the influence of the underlying approach on the result of the tool.
The selected computing tasks include training tasks of different nature (vision, language) and with different computational complexities, permitting us to evaluate the behaviour of the tested tools and methods in different settings. 

\emph{Contribution 3.} Our third contribution is open-source code both for the review and the experiments permitting not only to reproduce our results, but also to extend them.
Indeed, as we will see in Sections \ref{sec:protocol} and \ref{sec:protocol_execution}, the systematic review involves the search, selection, and classification of a large number of papers and we provide on \href{https://github.com/Accenture/Labs-Sustainable-AI/tree/slr_tools}{GitHub}\footnote{Full link: \href{https://github.com/Accenture/Labs-Sustainable-AI/tree/slr_tools}{\texttt{https://github.com/Accenture/Labs-Sustainable-AI/tree/slr\_tools}}} the repository containing the python scripts used to perform these steps (and the resulting data). By launching the scripts at a later time, it would be possible to enrich our systematic review with more recent papers.
Similarly, one can find on \href{https://github.com/Accenture/Labs-Sustainable-AI/tree/nrj_eval_comparison}{GitHub}\footnote{Full link: \href{https://github.com/Accenture/Labs-Sustainable-AI/tree/nrj_eval_comparison}{\texttt{https://github.com/Accenture/Labs-Sustainable-AI/tree/nrj\_eval\_comparison}}} the repository of the scripts used for the experiments; the repository may be extended with further ML computing tasks and energy evaluation tools and methods.




\subsection{Paper Overview and Outline}

This article is organized as follows. In Section \ref{sec:protocol} we define the protocol of the \acrshort{slr}, including how studies are collected (Section \ref{sec:collection}) and selected (Section \ref{sec:selection}), and how studies are classified and data is extracted from them (Section \ref{sec:classification_data_extraction}). Section \ref{sec:protocol_execution} details the actual execution of the protocol, notably the timeline and the number of papers identified during the search (and selection) steps. Then, in Section \ref{sec:summary}, we present the selected results, first introducing a taxonomy of the identified papers (Section \ref{sec:taxonomy}), then presenting the selected primary and secondary studies in Sections \ref{sec:presentation_primary_studies} and \ref{sec:presentation_secondary_studies}, respectively. Finally, we provide experimental results where a selection of tools is tested on different ML training tasks in Section \ref{sec:experiments}, and give our conclusions in Section \ref{sec:conclusion}.


\section{Protocol of the Review}
\label{sec:protocol}


In this section, we present the research protocol of this \acrshort{slr}, which is mainly based on the guidelines provided in \cite{kitchenham2007}. The protocol permits to built a \emph{pool} of items (scientific articles, reports, etc.) from which data is then extracted. We may divide the protocol in three main steps:
\begin{enumerate}
    \item collection of a pool of items,
    \item selection of the items,
    \item classification the selected items and data extraction.
\end{enumerate}
As we see in more detail in the following sections, we target items in which the authors have developed a specific method or tool, whether or not they have been specifically designed for ML applications, and we also target items in which the authors have tested methods and tools built by others.


\subsection{Collection of a Pool of Items}
\label{sec:collection}

\subsubsection{Data Sources}
\label{sec:data_sources}

To cover publications in the domains of Computer Science and Software Engineering, we use 3 data sources. On the one hand, we use the following two digital databases: \emph{ACM Digital Library}, an academic database for computer science, and \emph{IEEE Xplore Digital Library}, which covers journals and conference papers, technical standards, as well as some books, on electrical engineering, computer science, and electronics. We complement these data sources by the \emph{Google Scholar search engine} (GS). Indeed, GS covers a larger portion of the literature than most data sources, as well as much unpublished work across all scientific fields and in particular those of interest here, i.e., machine learning, computing, and energy.

Each of these data sources has its own specificity. The data source presenting most constraints for the systematic search process is GS. First, a search query may contain at most 256 characters. Second, for a given query, GS provides at most 1000 results even if the actual number of results associated to this query (which is displayed by GS) is greater. Finally, the GS official interface only permits the user to save results to the user's Google Scholar Library by selecting the \textit{star} icon, and then exporting said library. GS tends to block any behavior deviating from this usage mode. However SerpAPI permits to circumvent this issue (see \href{https://serpapi.com/}{\texttt{https://serpapi.com/}}). ACM permits to export results page by page (in BibTeX, EndNot or ACM Ref format), and the maximum number of results displayed on a single page is 50. IEEE permits to export all results at once (in csv format), though only the 2000 first results will be exported.

\subsubsection{Initial Pool} 
\label{sec:initial_pool}

The construction of our systematic review protocol notably builds upon an initial pool identified through some less structured search on the internet (whose results have been verified by a person) and on the basis of suggestions from experts in the domain.

This initial pool contains 13 items describing the development of a tool: 
Carbon-Tracker \cite{anthony2020}, 
Code-Carbon, previously developed under the name Energy-Usage \cite{lottick2019},
Deep-Neural-Network-Estimation-Tool \cite{yang2017},
Eco2AI \cite{budennyy2022},
ESAVE \cite{pathania2023},
Experiment-Impact-Tracker \cite{henderson2020},
PowerJoular and JoularJX \cite{noureddine2022},
Green Algorithms \cite{lannelongue2021},
LIKWID-powermeter \cite{treibig2010},
ML-CO2-Impact \cite{lacoste2019},
PMT \cite{corda2022},
PowerAPI \cite{bourdon2013}, 
Cumulator \cite{trebaol2020}.
The initial pool also contains 7 studies describing methods: \cite{rodrigues2018, rodrigues2020} (SyNERGY), \cite{shahid2021b}, \cite{shahid2021a}, \cite{singh2013}, \cite{strubell2019}, \cite{qiu2023} (for federated learning), and \cite{desislavov2023}. It contains as well 5 secondary studies (reviews): \cite{bannour2021}, \cite{garcia-martin2019a}, \cite{fahad2019}, \cite{noureddine2013}, and \cite{jay2023}.

We are also aware of four tools without any associated scientific study: Energy-Scopium, PyJoules, Perf, Scaphandre. References to such ``separate-tools'' that do not have any corresponding scientific study, are provided in specific tables, one located in Appendix \ref{ap:subsitiary_tools_alone_not_in_selection} for the four above tools, as well as Table \ref{tab:in_secondary_alone} for separate-tools discussed in Section \ref{sec:presentation_secondary_studies}.

This initial pool notably permits us to identify keywords associated with our research question (see Section \ref{sec:keywords}).

\subsubsection{Keywords} 
\label{sec:keywords}

To search for items, we look for results containing, in the title field, at least one word from each of the three following lists of keywords:
\begin{enumerate}[label=(\roman*)]
    \item machine learning, deep learning, computing, information and communications technology, \acrshort{ict}, artificial intelligence, AI, natural language processing, \acrshort{nlp}, neural network, neural networks, \acrshort{cnn}, DNN, computation, computations, software, process-level, server, virtual machine, federated learning, distributed learning;

    \item measure, measuring, estimate, estimation, consumed, consumption, predict, prediction, predicting, track, tracking, report, reports, reporting, account, quantify, quantifying, monitor, monitoring, evaluate, evaluating;

    \item energy, power, environmental impact, carbon footprint, carbon emissions, carbon impact.
\end{enumerate}
The first category of keywords initially also contained the words ``process'' and ``processes.'' The latter have finally been removed because they induced too many irrelevant results pertaining to industrial processes. As explained in Section \ref{sec:data_sources}, GS comes with a number of constraints. Mainly in view of reducing the number of results provided by GS, we simultaneously exclude all results containing any of the following keywords:
\begin{itemize}
    \item[(iv)] wind, building, buildings, vehicles, homes, ships, solar, photovoltaic, vehicle.
\end{itemize}
Here ``\acrshort{cnn}'', ``\acrshort{nlp}'' and ``DNN'' correspond to Convolutional Neural Network, Natural Language Processing and Deep Neural Network, respectively.

In the case of IEEE, we use an additional filter based on metadata. We select only the papers for which the ``Publication Topics'' contains at least one of the following values: “power consumption,” “energy consumption,” or “power aware computing”.


\subsubsection{Building Queries}
\label{sec:queries}

Each of the selected data sources allows for the use of the operators \textit{AND}, \textit{OR}, \textit{NOT}, \textit{parenthesis} and quotation to search for a specific phrase. In the first step, search by keywords, we use these operators and the keywords presented in Section \ref{sec:keywords}, to build appropriate queries for each of the data sources. The syntax of the queries differs slightly for the different data sources (see Appendix \ref{ap:queries}). In the case of GS, the original query is actually divided into a total of 103 sub-queries in order to meet the constraints of the data source (see Section \ref{sec:data_sources}). The results of these sub-queries are then merged together, removing the duplicates.

\subsection{Selection of the Items} \label{sec:selection}

\subsubsection{Selection Criteria}
\label{sec:selection_criteria}

We consider different dimensions for each item (relevance, type of literature, accessibility, language) and we include items which satisfy at least one inclusion criterion for each aspect. The inclusion criteria are as follows:
\begin{itemize}
    \item
\emph{Relevance to research question}: 
\begin{itemize}
\item include items where the authors develop tools or methods (not shaped into tools) that can measure, estimate or predict the energy consumption (or power profile) of machine learning (training and/or inference); if the tool/method has not been specifically designed for machine learning, but rather for a broader range of computing tasks, the item should still be included;
\item include items where tools or methods are being used for measuring/estimating the energy consumed by machine learning, even though not developed by the authors;
\end{itemize}

\item\emph{Literature type}:
\begin{itemize}
\item include ``\emph{articles}'': peer-reviewed scientific articles (journals, conferences, workshops), or parts of books;
\item include ``\emph{preprints}'': non-peer reviewed scientific articles, which may for instance be found on arXiv, Reasearch Gate, Hal, as well as on the author's personal website;
\item include other materials and research produced outside of the traditional commercial or academic publishing and distribution channels, such as technical reports, thesis, white papers, etc.;
\end{itemize}

\item\emph{Accessibility}: only include items for which the full text is available;

\item\emph{Language}: only include items written in English.
\end{itemize}

\subsubsection{Semi-Automatic Selection}
\label{sec:semi_automatic_selection}
Our first selection step is a semi-automated selection phase. This phase is based on the results' titles and on the relevance to the research question only. It consists in identifying words (among all words contained in the results' titles) that rule a title containing any of these words, as off-topic. Indeed, our query captures studies on energy efficiency in the domain of renewable energies, manufacturing, construction, etc. However, many of these studies' titles contain common words that permit to easily identify them as off-topic with respect to our research question. 
The list of words and pairs of words to exclude from the titles is provided in Appendix \ref{ap:excluding_words}.

\subsubsection{Selection by Hand}
\label{sec:selection_by_hand}
The second part of the selection process is based on assessors reading the results' titles, abstracts and full text if necessary. 
Three assessors share this task. Doubts on a result's selection are resolved through discussion with another assessor.
We divide the selection by hand in two parts. The first part is solely based on the selection criteria described in the Section \ref{sec:selection_criteria}. 

The second part is based on additional selection criteria that have been added during the protocol application. The aim of this second part of the selection by hand, \textcolor{\colorA}{is} to reduce the size pool of selected items, by excluding items least relevant to our research question.
Here, we additionally exclude items for which both:
\begin{itemize}
    \item the authors have not created a method or tool, but rather used one created by others,
    \item the authors have not tested this method or tool on ML applications, but rather on other computing tasks.
\end{itemize}

\subsection{Classification of the Selected Items and Data Extraction}
\label{sec:classification_data_extraction}

\subsubsection{Classification}

In the fourth step, we classify the results selected in the third step according to two criteria:
\begin{enumerate}
    \item First, is the result a primary study or a survey? Then, if the result is a primary study, has a tool or method been created by the authors or not? (The latter case implies a tool or method is used by the authors.) We provide one of the three following values: ``Yes -- creation,'' ``No -- no creation,'' ``Survey.''
    \item Is the result specifically concerned with ML applications or not? (In the latter case the result is concerned with software in general, virtual machines, data centers, etc.) We provide one of the two following values: ``Yes -- for ML,'' ``No -- not for ML.''
\end{enumerate}
This permits to obtain six groups of studies, as presented in Table \ref{tab:classification_groups}: $\mathscr{{Y}Y}$ (creation \& for ML), $\mathscr{{N}Y}$ (no creation \& for ML), $\mathscr{{Y}N}$ (creation \& not for ML), $\mathscr{{N}N}$ (no creation \& no for ML), $\mathscr{{S}Y}$ (survey \& for ML) and $\mathscr{{S}N}$ (survey \& not for ML). 
Note that the group $\mathscr{{N}N}$ would contain the items excluded in the second part of the selection by hand, described in Section \ref{sec:selection_by_hand}, as it concerns studies where there is no tool creation and no application to ML by the authors.

\begin{table}
    \centering
    \begin{tabular}{|l|l|l|}
      \hline
      \diagbox{creation}{for ML} & yes & no\\
      \hline
      yes & $\mathscr{{Y}Y}$ & $\mathscr{{Y}N}$ \\
      \hline
      no & $\mathscr{{N}Y}$ & $\mathscr{{N}N}$ \\
      \hline
      survey & $\mathscr{{S}Y}$ & $\mathscr{{S}N}$ \\
      \hline
    \end{tabular}
    \caption{Classification of selected studies in six groups.}
    \label{tab:classification_groups}
\end{table}

\subsubsection{Data Extraction Forms}


We prepare data extraction forms for all five groups of items described in \ref{sec:classification_data_extraction}: $\mathscr{{Y}Y}$, $\mathscr{{N}Y}$, $\mathscr{{Y}N}$, $\mathscr{{S}N}$, and $\mathscr{{S}Y}$. The results of these extractions are presented in Section \ref{sec:summary}.
Lets us start with primary studies.

\paragraph*{Group $\mathscr{{Y}Y}$.} For items with tool creation and applications to ML we ask questions belonging to six categories that we call ``study,'' ``detail,'' ``target task,'' ``constraints,'' ``available,'' and ``cites.'' The questions are the following.
\begin{itemize}
    \item Study: 
    \begin{itemize}
        \item Provide the name of the tool or method, or ``Name Unspecified'' (\acrshort{nu}) for tools and methods that are unnamed.
        \item Provide the reference of the item.
        \item Provide the publication or appearance year of the item.
    \end{itemize}
    \item Detail: 
    \begin{itemize}
        \item What is the approach behind the method or tool developed by the authors (e.g., estimation model, sensors, measurements, etc.)? If it is an estimation model, provide detail on the type of model and inputs to the models.
        \item What part of the computing task is accounted for/targeted by the method or tool (e.g., data movement, computations)?
        \item Does the method or tool account for the energy consumption of specific hardware? If yes, which hardware?
    \end{itemize}
    \item Target Task: What resources are accessible to us in connection with this tool or method (such as code, models, APIs)?
    \item Constraints: If any, what are the hardware or software constraints for using this method or tool?
    \item Available: What is available to us related to this tool or method (code, model, API, etc.)?
    \item Cites: How many citations does the item have according to Google Scholar?
\end{itemize}

\paragraph*{Group $\mathscr{{N}Y}$.} For items without tool creation and with applications to ML we ask questions belonging to four categories that we call ``study,'' ``detail,'' ``ML task,'' and ``setup''. The questions are the following.
\begin{itemize}
    \item Study:
    \begin{itemize}
        \item Provide the reference of the item.
        \item Provide the publication or appearance year of the item.
    \end{itemize}
    \item Detail: 
        \begin{itemize}
        \item  Provide the name or type of method or tool used by the authors.
        \item If the latter tool or method also belongs to the pool of selected items, provide its reference.
    \end{itemize}
    \item ML Task: On what ML task is the method or tool being used by the authors?
    \item Setup: What is the hardware setup of the authors?
\end{itemize}

\paragraph*{Group $\mathscr{{Y}N}$.} For items with tool creation and no applications to ML we ask questions belonging to three categories that we call ``study,'' ``detail,'' and ``cites.'' The questions are the following.
\begin{itemize}
    \item Study: 
    \begin{itemize}
        \item Provide the name of the tool or method, or ``Name Unspecified'' (\acrshort{nu}) for tools and methods that are unnamed.
        \item Provide the reference of the item.
        \item Provide the publication or appearance year of the item.
    \end{itemize}
    \item Detail:
    \begin{itemize}
        \item What is the approach behind the method or tool developed by the authors (e.g., estimation model, sensors, measurements, etc.)? If it is an estimation model, provide detail on the type of model and inputs to the models.
        \item What part of the computing task is accounted for/targeted by the method or tool (e.g., data movement, computations)?
        \item Does the method or tool account for the energy consumption of specific hardware? If yes, which hardware?
        \item What is available to us related to this tool or method (code, model, API, etc.)?
    \end{itemize}
    \item Cites: How many citations does the item have according to Google Scholar?
\end{itemize}

\paragraph*{Groups $\mathscr{{S}Y}$ and $\mathscr{{S}N}$.}
Finally, for secondary studies we ask questions belonging to three categories that we call ``context,'' ``taxonomy,'' and ``content''. The questions are the following.
\begin{itemize}
    \item Context: What was the aim of the authors?
    \item Taxonomy: If any, what type of taxomony or categorisation do the authors use in their review?
    \item Content: Provide a list or overview of the tools and methods reviewed by the authors.
\end{itemize}
For secondary studies, we record the number of tools and methods reviewed by the authors. We also record tools that are without any associated literature (where by ``literature'', we mean the ``type of literature'' described in the selection criteria in Section \ref{sec:selection_criteria}) but are cited in these reviews. We call such tools ``\emph{separate-tools}.''

\section{Execution of the Protocol}
\label{sec:protocol_execution}



\begin{figure}
    \centering
    \includegraphics[width=12cm]{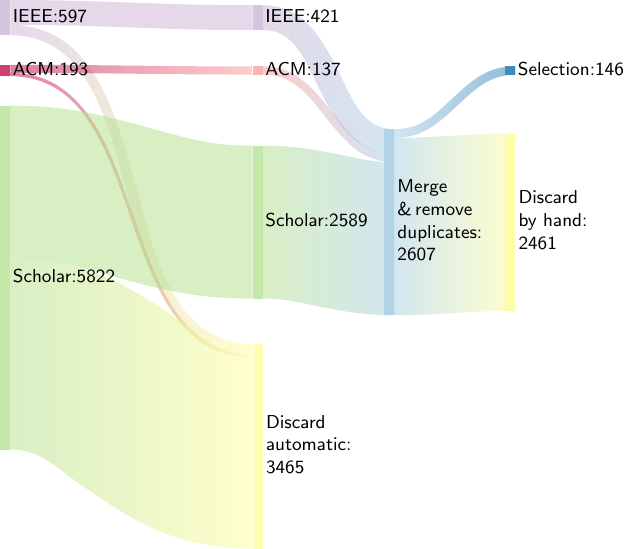}
    \caption{Caption: execution of the protocol -- part I}
    \label{fig:sankey_SLR_phase1}
\end{figure}

\begin{figure}
    \centering
    \includegraphics[width=12.5cm]{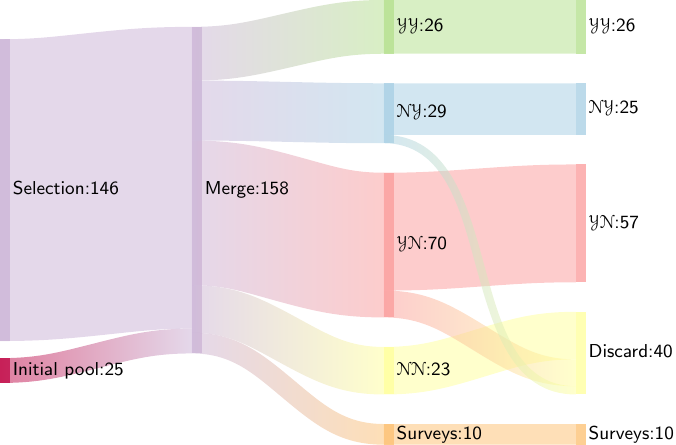}
    \caption{Caption: execution of the protocol -- part II}
    \label{fig:sankey_SLR_phase2}
\end{figure}

\begin{figure}
    \centering
    \begin{tikzpicture} 
    \draw[->, thick] (0, 0) -- (14,0); 
    \draw[shift={(0,0)}, color=black] (0pt,2pt) -- (0pt,-2pt); 
    \node[below] at (0, 0) {06/23};
    
    \draw[shift={(3,0)}, color=black] (0pt,2pt) -- (0pt,-2pt); 
    \node[below] at (3, 0) {07/23}; 
    
    \draw[shift={(6,0)}, color=black] (0pt,2pt) -- (0pt,-2pt); 
    \node[below] at (6, 0) {08/23}; 
    
    \draw[shift={(9,0)}, color=black] (0pt,2pt) -- (0pt,-2pt); 
    \node[below] at (9, 0) {09/23};
    
    \draw[shift={(12,0)}, color=black] (0pt,2pt) -- (0pt,-2pt); 
    \node[below] at (12, 0) {10/23};
    
    \node[above, align=center] at (12/10,0) {12/06/23 \\ Search \vspace{2mm}};
    \draw[shift={(12/10,0)}, color=black, ultra thick] (0pt,2pt) -- (0pt,-2pt);
    
    \node[above, align=center] at (3+20/10,0) {20/07/23 \\ Semi-automatic selection \vspace{2mm}}; 
    \draw[shift={(3+20/10,0)}, color=black, ultra thick] (0pt,2pt) -- (0pt,-2pt);
    
    \node[above, align=center] at (9+7/10,0) {07/09/23 \\ Classification \vspace{2mm}}; 
    \draw[shift={(9+7/10,0)}, color=black, ultra thick] (0pt,2pt) -- (0pt,-2pt);
    
    \node[above, align=center] at (12+6/10,0) {06/10/23 \\ Data extraction \vspace{2mm}}; 
    \draw[shift={(12+6/10,0)}, color=black, ultra thick] (0pt,2pt) -- (0pt,-2pt);
    \end{tikzpicture}
    \caption{Caption: Timeline of the search steps}
    \label{fig:protocol_timeline}
\end{figure}
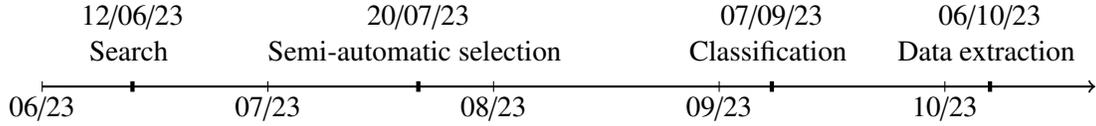

Let us now describe the execution of the protocol (see Figure \ref{fig:protocol_timeline} for the associated timeline) and the proportion of results across each search step. The initial search through the IEEE Digital Library, ACM Digital Library, and Google Scholar yielded 597, 193, and 5822 results, respectively. We thus obtained an initial set of 6612 results. Proceeding with the semi-automatic selection phase described in Section \ref{sec:semi_automatic_selection} with a total of 688 ``excluding words,'' we discarded a first set of 3465 results, obtaining a smaller set of 3147 results: 421 from IEEE, 137 from ACM and 2589 from GS. We then merged these three groups and removed all duplicates, obtaining a total of 2607 results. Finally, the first part of the selection by hand yielded 146 selected results (and 2607 additional discarded results). We refer to Figure \ref{fig:sankey_SLR_phase1} for visualization. Merging this selection with our initial pool of papers (see Section \ref{sec:initial_pool}), we added 12 additional studies to obtain a total of 158 results. 
We then proceeded with the second part of the selection by hand step (described at the end of Section \ref{sec:selection_by_hand}) and obtained, after discarding 41 additional results (including 23 from the group $\mathscr{{N}N}$ as explained in Section \ref{sec:selection_by_hand}), the following final selection of 118 studies: 26 in $\mathscr{{Y}Y}$, 25 in $\mathscr{{N}Y}$, 57 in $\mathscr{{Y}N}$, 3 in $\mathscr{{S}Y}$ and 7 in $\mathscr{{S}N}$.

\section{Overview and Summary of the Selected Items}
\label{sec:summary}


We may now describe the 118 selected results. We describe each of the five groups separately ($\mathscr{{Y}Y}$, $\mathscr{{N}Y}$, $\mathscr{{Y}N}$, $\mathscr{{S}Y}$, $\mathscr{{S}N}$), starting with the secondary studies in Section \ref{sec:presentation_secondary_studies}, before going to the primary studies in Section \ref{sec:presentation_primary_studies}. For the latter, we also compare studies in terms of the approach used by the method/tool to evaluate energy consumption. In view of this we have chosen the taxonomy described in the following Section.

\subsection{Taxonomy}
\label{sec:taxonomy}

\begin{figure}
    \centering
    \includegraphics[width=10cm]{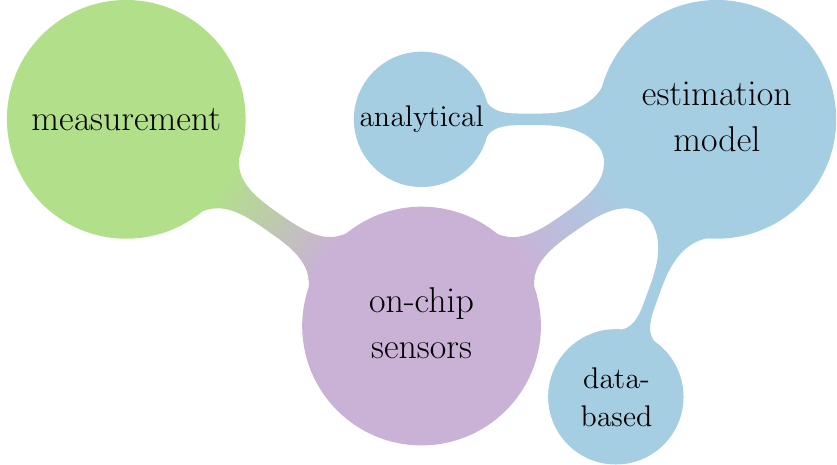}
    \caption{Taxonomy of energy evaluation techniques for computing tasks.}
    \label{fig:taxonomy}
\end{figure}

We count four different categories of approaches (techniques) to evaluate energy consumption of ML tasks or general computing tasks (see Figure \ref{fig:taxonomy}): measurement, data-based estimation model, analytical estimation model, and on-chip sensors:
\begin{itemize}
    \item Measurement: \acrfull{epm} or sensor measure power, current intensity and/or voltage, for the whole computer or specific hardware components.
    \item Estimation model: an estimation model takes indirect evidences, such as activity factors (in other words, useful features, hardware or software provided metrics) or characteristics of the target application (e.g. a neural network's architecture) and matches them to an energy consumption or power draw output. We differentiate between two kinds of models:
    \begin{itemize}
        \item Data-based model: the model learns patterns and relationships directly from a data set through algorithms such as a ML model or a statistical model.
        \item Analytical model: explicit equations or formula describe relationships between variables.
    \end{itemize}
    \item On-chip sensors based approaches: on-chip sensors, such as \acrfull{rapl} and \acrfull{nvml}, are sensors embedded in some vendors' processors with associated libraries to access their data. Although such approaches may overlap with the measurement and estimation groups (see Section \ref{sec:on-chip_sensors} for detail), we consider them as a separate group, as they are based on vendor specific tools.
\end{itemize}

Some approaches may target a single hardware component, such as the CPU or the GPU. Others may aggregate the energy consumption of multiple hardware components, intrinsically, as is the case of an external power meter placed at the power outlet level, or artificially, by summing, for instance, the consumption provided (by one or several of the above approaches) for several hardware components. A simple summation may be replaced by a modelling approach itself. Indeed, the aggregation may be done by means of an analytical modelling (such as a simple weighted sum with predetermined coefficients) or by means of a data-based model (for instance learning from data the coefficients representing the contribution of the different hardware components to the total energy consumption of the system).

Some of the studies reviewed here belong to several of the four categories. This is particularly common for approaches aggregating the consumption of multiple hardware components. First, the aggregation method can involve using analytical (and even sometimes data-based) modelling to some degree. Second, some techniques may combine different approaches for different components, for instance an analytical estimation model for the CPU with an on-chip sensor approach for the GPU. Furthermore, even an approach targeting a single hardware component may be a combination of several of the four categories.



\subsubsection{Measurement}
As already mentioned above, the measurement category refers to actual measurements of current, power, voltage. These measurements can be done at different places, ranging from a wall outlet to measurements at the motherboard (\cite{mobius2014} notably reviews several measurement approaches), thus requiring additional or specialised hardware such as a multi-meter or a specialized circuit integrated into the motherboard \cite{noureddine2013, noureddine2015}. This approach is considered to be the baseline (ground truth) for energy consumption evaluation. However, at large-scale, utilizing \acrshort{epm}s becomes costly \cite{alavani2023}. Moreover, a drawback of this approach that is recurrently pointed out \cite{fahad2019, noureddine2013, shahid2021a}, is its inability to furnish fine-grained decomposition of the energy consumption. An \acrshort{epm} placed at the power outlet cannot provide information of where the power is consumed in the computer and even specialized integrated circuits with power sensors cannot monitor the consumption of a specific software (and even less, its classes and methods' usage) \cite{noureddine2013, noureddine2015, shahid2021a, mobius2014}.
In order to circumvent this issue and use \acrshort{epm} as a baseline for more fine-grained energy evaluation methods and tools, some studies, such as \cite{hankel2016}, \cite{fahad2019} and \cite{shahid2021a} have proposed specific experimental settings. For example in \cite{hankel2016} all the variables that can impact variation in energy consumption (e.g., fans) are fixed so as to target a specific software performing a function. Similarly, in \cite{shahid2021a} the authors ensure that the value of dynamic energy is only due to the CPU and RAM.
Furthermore, in \cite{jay2023} the authors compare Energy Scopium, Scaphandre, Perf (see Table \ref{tab:in_secondary_alone}) and PowerAPI \cite{bourdon2013}, a series of fine-grained energy evaluation methods that provide power profiles (power evolution through time), with \acrshort{epm} and Baseboard Management Controllers (measurement equipment placed inside computing nodes), in terms of correlations. Overall, they observe similar and strongly correlated power profiles, and some differences are analyzed and discussed in detail by the authors.

\subsubsection{Inputs of the Estimation Models}
\label{sec:estimation_model_inputs}

The analytical and data-based estimation models' inputs (or predictor variables) can range from Performance Monitoring Counters (\acrshort{pmc}s), and hardware utilization rates, to hardware specifications, and software characteristics such as the architecture of the \acrfull{nn} to be trained/executed, or \acrfull{ptx} code (an intermediate compilation of CUDA code generated at compile time).

\acrshort{pmc}s are registers provided in the processor to store the counts of software and hardware activities \cite{fahad2019, shahid2021b}. This information, collected during programs' execution, sheds light on the behavior of these programs. 
\acrshort{pmc}s are thus metrics directly provided by the hardware, as opposed to OS provided metrics (i.e., computed by the OS) such as the utilization level of a system (e.g., the CPU utilization indicator) \cite{mobius2014}.
\acrshort{pmc} are also referred to as ``hardware performance counters'' \cite{schubert2012}, ``power monitoring units'' \cite{gutierrez2015b}, ``performance events'' \cite{fu2018} in the literature reviewed here.



In addition to aforementioned inputs, other types of inputs observed in the reviewed studies include: 
\begin{itemize}
    \item characteristics of \acrshort{nn} layers such as their shape, number of non-zero values and bitwidths (i.e., bits used to represent each value in a numerical data type);
    \item hardware specifications (constant parameters associated to a specific hardware and generally provided by the manufacturer) such as the \acrfull{tdp} of the CPU or GPU, which is the maximum heat flow generated by a CPU or GPU that its cooling system is designed to dissipate; the TDP can be seen as an indication of the maximum power the component can draw;
    \item execution and memory access traces obtained by means of simulators,
    \item static features of sources code (obtained without executing the program), compiled binary;
    \item Floating Point Operations (\acrshort{flop}s) or \acrfull{mac} count;
    \item task duration.
\end{itemize}


\subsubsection{Data-Based Estimation Models}


As explained in \cite{mobius2014}, data-based estimation models (referred to as ``power estimation models'' in \cite{mobius2014}) are typically built via two essential steps: first, the selection of the model's inputs (see Section \ref{sec:estimation_model_inputs}), and second, the identification of a tool to train and test the model. The latter may be a benchmark, that the authors define as ``software programs specifically designed to stress some of the subsystems of a server in a comprehensible and repeatable manner.'' According to \cite{fahad2019}, a model is typically trained using a large suite of diverse benchmarks and validated against a subset of the benchmark suite and some real-life applications.
%
%
The estimation error is then the difference between the estimated power consumption and the actual power consumption, also called baseline, ground truth or reference. The way to quantify the baseline varies across studies, some of them, e.g. \cite{fu2018}, use on-chip sensors (see Section \ref{sec:on-chip_sensors}), and others, e.g. \cite{ma2009}, use actual measurements.
As reported by \cite{mobius2014}, the estimation error is significantly influenced by the choice of 1) the model input parameters, 2) the model training techniques, 3) the benchmarks/applications for training and evaluation purposes, and 4) the power baseline to which the estimated power is compared.
%
Several studies agree that the most common approach used to build a data-based estimation model for energy consumption is linear regression \cite{mobius2014, fahad2019, shahid2021a, shahid2021b, gutierrez2015a}. Some also state that models inputs are generally \acrshort{pmc}s recorded at the target hardware components during a program's execution. Often, one model is built for each component and the consumption of each component is then summed \cite{fahad2019}. In \cite{gutierrez2015a} estimation models based on \acrshort{pmc}s are notably reviewed. 

While the \acrshort{pmc}s and other activity factors are typically recorded during an application run, one can also use simulators (emulating a specific hardware platform and integrating monitoring tools into the code whose execution is simulated on that platform) to obtain these counters and thus bypass the need to execute an application \cite{alavani2023}. In \cite {garcia-martin2019a}, the authors look at both approaches that simulate hardware, and approaches that monitor \acrshort{pmc}s, and discuss their respective advantages and disadvantages. They notably observe that, while simulation approaches provide detailed results but with a significant overhead, PMC approaches have no overhead but cannot provide per-processor results.

%
%
In \cite{mobius2014, alavani2023}, the authors also note that \acrshort{pmc}s are architecture-specific and the estimation models may thus not transfer well from one architecture to another. 
Indeed, some studies explicitly mention that their model need to be trained on the computer they are to be used on, explaining that a so-called ``calibration phase'' is needed before using the model \cite{schubert2012, lewis2008}. However, \cite{gutierrez2015a} observes that training models on one machine and applying them to another with a significantly different architecture may yield acceptable results, and the authors thus suggest that models may be directly transferable when applied to machines with similar architectures.


An example of how a model can be built and in particular how data for training (software performance data against energy consumption) can be gathered, is found for instance in \cite{fu2018} where the authors make use of the Perf tool (see Table \ref{tab:in_secondary_alone}) and detail all the steps, including energy consumption data collection, correlation analysis between performance features and energy consumption features, feature selection, and the selection of ML algorithms to model energy consumption.
Concerned with estimation models for the CPU, \cite{mobius2014} notes that most models require knowledge of the architecture of the CPU and the nature of the benchmark/application to select appropriate \acrshort{pmc}s. 
In \cite{shahid2021a}, the authors review notable estimation models based on \acrshort{pmc}s and observe that despite the advantages of this approach in terms of its cost and fine-grained nature (compared to \acrshort{epm}s), the construction of such model presents several issues. Indeed, they state that model construction is complex, complaining that ``the majority of research works select \acrshort{pmc}s solely based on their high positive correlation with energy consumption without any deep understanding of the model variables’ physical significance,'' and that there is a ``lack of consensus among the research works, reporting prediction accuracy ranging from poor to excellent'' as well as a lack of understanding of the causes of the estimations inaccuracy. To address such issues, the authors propose a theory of modelling based on \acrshort{pmc}s. In particular, they make explicit the assumptions behind such models, formulating them in a mathematical form, and also extend the formalism by adding properties heretofore ignored. The practical implications of their theory notably include selection criteria for models inputs and coefficients.


\subsubsection{Analytical Estimation Models}

A typical example of analytical estimation model is for instance computing the energy consumption of the CPU (or GPU) by the product of its \acrshort{tdp} and the total execution time of the target computing task \cite{lacoste2019}. This assumes that the CPU (or GPU) is utilized at 100\%, or in any case at some known constant average utilization level as in \cite{lannelongue2021}. 
The authors in \cite{noureddine2014a} also propose a variation of this model, in which the CPU \acrshort{tdp} is first multiplied by 0.7 to account for the fact that the actual power draw of the CPU is generally less than the amount of heat the component generates, that is stipulated by the \acrshort{tdp}.
Other models have been proposed that involve for instance \acrshort{flop} or \acrshort{mac} count, characteristics of the application (such as the architecture of a \acrshort{nn} in the context of a ML computing task), static features of source code, and \acrshort{pmc}s, see for instance \cite{desislavov2023}, \cite{lemaire2022}, \cite{liu2017}, and \cite{peng2013}, respectively.

\subsubsection{On-Chip Sensors}
\label{sec:on-chip_sensors}

Approaches based on on-chip sensors, also called internal interfaces \cite{jay2023}, rely on 1) sensors embedded in mainstream processors such as Intel and AMD Multicore CPUs, Nvidia GPUs, and Intel Xeon Phis and 2) associated vendor specific libraries that give access to power data from these sensors. Well known examples include \acrfull{rapl} for Intel CPUs, and \acrfull{nvml} for Nvidia GPUs and \acrfull{smc} for Intel Xeon Phi \cite{shahid2021a} (see Tables \ref{tab:in_secondary_not_alone} and \ref{tab:in_secondary_alone}).
In particular, \acrshort{rapl} may report energy consumption of the CPU at different levels: entire CPU socket ``PKG,'' all CPU cores ``PP0,'' integrated graphics ``PP1,'' dynamic random-access memory ``DRAM,'' and entire SoC ``PSys'' \cite{jay2023}. 

Several studies report a lack of information about on-chip sensors' energy consumption evaluation methodology and implementation detail \cite{jay2023, alavani2023, fahad2019, shahid2021a}, and the consequent lack of knowledge about their accuracy (apart for \acrshort{nvml} \cite{shahid2021a}). 
It is even unclear whether some on-chip sensors provide actual measurements or estimation by means of models (based on \acrshort{pmc}s for instance).

For instance, the latest version of \acrshort{rapl} is said to involve voltage regulators. According to \cite{fahad2019}, these voltage regulators keep track of an estimate of the current, without much information about the underlying estimation method. Furthermore, \cite{fahad2019} states that \acrshort{rapl} predicts the energy consumption of CPUs and RAM based on an undisclosed set of \acrshort{pmc}s.
On the contrary, \cite{jay2023} claims that, while the first version of \acrshort{rapl} relies on an estimation model based on ``a set of architectural events from each Intel architecture core, the processor graphics, and I/O,'' the new version of \acrshort{rapl} enables actual power measurement, with significantly more accurate estimates. Besides, recent studies remark the need for administrator access to utilize \acrshort{rapl} \cite{bannour2021, caspart2023}.

Similar questions arise for \acrshort{nvml}, with no clear indication of whether the power is measured directly or estimated~ \cite{jay2023}. However, \cite{alavani2023}~categorizes \acrshort{nvml} as a ``direct method,'' in contrast to ``indirect methods'' such as estimation models and simulators.

\subsection{Summary of the Selected Methods and Tools}
\label{sec:presentation_primary_studies}

We will now summarize, by means of tables, the 108 selected primary studies, presenting successively each of the groups $\mathscr{{Y}Y}$, $\mathscr{{N}Y}$ and $\mathscr{{Y}N}$ in Sections \ref{sec:presentation_YY}, \ref{sec:presentation_NY} and \ref{sec:presentation_YN}, respectively. In each section, we group studies according to the taxonomy described in Section \ref{sec:taxonomy}. The column ``study'' contains the reference of the article, year of publication and name of the tool or method created by the authors (for the groups $\mathscr{{Y}Y}$ and $\mathscr{{Y}N}$).

In what follows, we use the subsequent abbreviations: ``\acrshort{nn}'' for Neural Network, and ``\acrshort{nu}'' for Name Unspecified (for tools and methods that are unnamed). Whenever code or trained models have been made available by the authors, we provide the corresponding link directly in the tables, and one can also refer to Appendix \ref{ap:links} for the list of URLs.

\subsubsection{Studies with Tool Creation and Applied to ML ($\mathscr{{Y}Y}$ group)}
\label{sec:presentation_YY}

\begin{figure}[h!]
    \centering
    \input{tables/YY-timeline}
    \caption{Timeline of the studies $\mathscr{{Y}Y}$.}
    \label{fig:timeline_YY}
\end{figure}
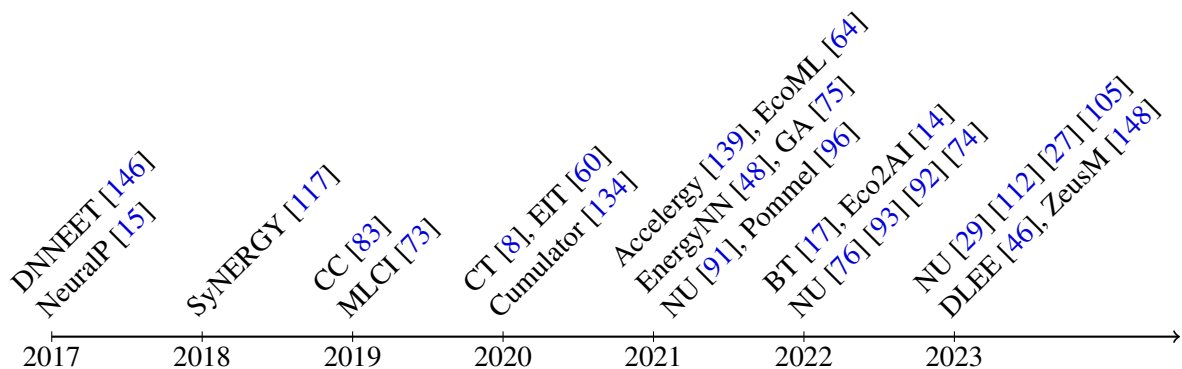

The studies of the group $\mathscr{{Y}Y}$ (see Figure \ref{fig:timeline_YY} for the associated timeline) are reported in six different tables: in Section \input{tables/YY-description} We detail 1) which type of meter, sensors or model has been developed by the authors and for what part of the application, system and/or hardware (column ``detail''), and 2) which computing task the tool or method may work for (column ``target task''). Whether there are any specific known hardware or software constraints to the method or tool, and whether any package, code, trained model or API (or website) is available, is indicated in the columns ``constraints'' and ``available,'' respectively. Finally, the column ``cites'' contains the number of citations of the corresponding scientific article (reported by Google Scholar).


\input{tables/YY-analytical-estimation-model}
\input{tables/YY-data-based-estimation-model}
\input{tables/YY-on-chip-sensors}
\input{tables/YY-analytical-and-data-based-estimation-model}
\input{tables/YY-analytical-estimation-model-and-on-chip-sensors}
\input{tables/YY-other}


\subsubsection{Studies with Tool Usage and for ML ($\mathscr{{N}Y}$ group)}
\label{sec:presentation_NY}

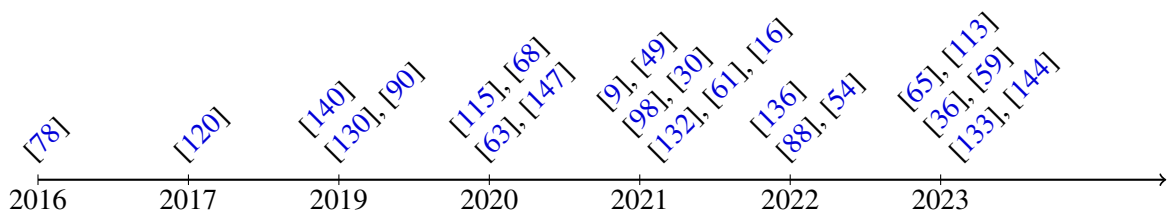
\begin{figure}[h!]
    \centering
    \input{tables/NY-timeline}
    \caption{Timeline of the studies $\mathscr{{N}Y}$.}
    \label{fig:timeline_NY}
\end{figure}

Here, for each study of the group $\mathscr{{N}Y}$ (see Figure \ref{fig:timeline_NY}), we detail the meter, sensors or model used by the authors (column ``detail''), on what ML task the latter have been used (column ``ML task''), and the setup or hardware context in which they have been used (column ``setup''). The studies are grouped in six different tables: in Section \input{tables/NY-description}


\input{tables/NY-measurement}
\input{tables/NY-analytical-estimation-model}
\input{tables/NY-on-chip-sensors}
\input{tables/NY-data-based-estimation-model}
\input{tables/NY-other}


\subsubsection{Studies with Tool Creation and not Applied to ML ($\mathscr{{Y}N}$ group)}
\label{sec:presentation_YN}

\begin{figure}[h!]
    \centering
    \input{tables/YN-timeline}
    \caption{Timeline of the studies $\mathscr{{Y}N}$.}
    \label{fig:timeline_YN}
\end{figure}

Finally, for the group $\mathscr{{Y}N}$ (see Figure \ref{fig:timeline_YN} for the associated timeline), we explain which type of meter, sensors or model has been developed by the authors and for what part of the application, system and/or hardware in the column ``detail'', as done for the group $\mathscr{{Y}Y}$. Moreover, if any package, code, trained model or API (or website) is available, we add the corresponding link. The studies are grouped in six different tables: in Section \input{tables/YN-description} For the mixed group, we additionally indicate the different approaches on which the method or tool is based.



\input{tables/YN-measurement}
\input{tables/YN-analytical-estimation-model}
\input{tables/YN-data-based-estimation-model}
\input{tables/YN-on-chip-sensors}
\input{tables/YN-hybrid}
\input{tables/YN-other}


\subsection{Summary of Selected Surveys}
\label{sec:presentation_secondary_studies}

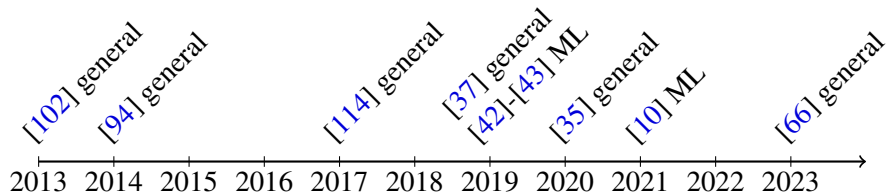
\begin{figure}[h]
    \centering
        \begin{tikzpicture}
        \draw[->, thick] (0,0) -- (11,0); 
        \foreach \x/\year in {0/2013, 1/2014, 2/2015, 3/2016, 4/2017, 5/2018, 6/2019, 7/2020, 8/2021, 9/2022, 10/2023} { 
            \draw[shift={(\x,0)}, color=black] (0pt,2pt) -- (0pt,-2pt); 
            \node[below] at (\x, 0) {\year}; 
            } 
        \node[above right, align=center, rotate=45] at (0,0) {\cite{noureddine2013} general}; 
        \node[above right, align=center, rotate=45] at (1,0) {\cite{mobius2014} general};
        \node[above right, align=center, rotate=45] at (4,0) {\cite{rieger2017} general};
        \node[above right, align=left, rotate=45] at (5.9,0) {\cite{fahad2019} general\\ \cite{garcia-martin2019b}-\cite{garcia-martin2019a} ML};
        \node[above right, align=center, rotate=45] at (7,0) {\cite{ergasheva2020} general};
        \node[above right, align=center, rotate=45] at (8,0) {\cite{bannour2021} ML};
        \node[above right, align=center, rotate=45] at (10,0) {\cite{jay2023} general};
        \end{tikzpicture}
    \caption{Timeline of the selected surveys.}
    \label{fig:timeline_surveys}
\end{figure}

We now summarize the 10 selected surveys (see Figure \ref{fig:timeline_surveys}), starting with surveys that are not focused on ML computing tasks. All tools or methods that are not part of the selected primary studies are listed in Table \ref{tab:in_secondary_not_alone} if they are associated with a scientific article, and in Table \ref{tab:in_secondary_alone} otherwise.

\subsubsection{Surveys not Focused on ML}

\emph{Jay and al. 2023} \cite{jay2023}. The authors perform an extensive experimental comparison of various software packages aimed at evaluating software energy consumption. The authors categorized these tools into \acrshort{epm}s, intra-node devices (sited between the power supply and the main board, capable of providing component-level information within the computing node), hardware sensors/software interfaces (such as on-chip sensors), and power and energy modeling. The reviewed tools are referred to as ``software-based power meters,'' grouped into three classes: ``energy calculators,'' ``energy measurement software,'' and ``power profiling software,'' where energy calculators estimate energy consumption using \acrshort{tdp}-based modeling, while energy measurement and power profiling software can respectively report total energy consumption or power draw over time for CPU, DRAM, and/or GPU (based on on-chip sensors). The assessed tools are: ML-CO2-IMPACT \cite{lacoste2019}, Green-Algorithm \cite{lannelongue2021} (energy calculators),  Carbon-Tracker \cite{anthony2020}, Code-Carbon \cite{lottick2019}, Experiment-Impact-Tracker \cite{henderson2020} (energy measurement software), and PowerAPI \cite{bourdon2013}, Energy-Scope (power profiling software), Perf, Scaphandre (see Table \ref{tab:in_secondary_alone}). The tools are compared on the basis of available features, supported sampling rates, and quality of estimation, and recommendations are provided on the most fitting tool to utilize based on distinct circumstances. Finally, while the tools were crafted for ML computing tasks, they are not tested for these applications in \cite{jay2023}. The authors have expressed their intention to carry out this more specific evaluation in the future.

\emph{Pijnacker and al. 2023} \cite{pijnacker2023}. The authors aim to enhance software's power efficiency to mitigate its environmental impact. In view of this, they search for tools able to measure and monitor the energy efficiency of software. Their analysis entails a rapid review encompassing 21 papers, including the following tools: SKD4ED \cite{marantos2021}, Energy-Toolbox \cite{marantos2022}, eCalc \cite{hao2012}, eTune \cite{ge2012}, Green-JEXT \cite{godboley2015}, GreenOracle \cite{chowdhury2016}, PowerScope \cite{flinn1999}, eProf \cite{schubert2012}, GreenSoM \cite{cordero2015}, FEETINGS/EET \cite{mancebo2018}, SPELL \cite{pereira2017}, SEFLab \cite{ferreira2013}, and Orka \cite{westfield2016}. They categorize the results based on target (e.g., java, general software, smartphones, data centers/cloud, etc.), granularity (ranging from system-level to procedure/method-level), and technique (e.g., whether the tool relies on hardware or software, and whether it involves estimation). The authors highlight that most tools primarily focus on general software systems and smartphone applications, targeting either the application- or class/component-level.
They note the absence of dedicated tools for commonly used software, and explain that the reviewed tools often apply to very specific software or hardware. 
Moreover, they observe that many estimation techniques, although often providing a reasonably accurate depiction of power usage and potential software enhancements, exhibit an error of about 10\% which may not be sufficient for some specific applications. They also raise the concern that measurement tools may necessitate specialized hardware, rendering them impractical or inaccessible for many software developers.
Overall, they conclude the necessity for a comprehensive set of tools that are widely applicable (wider range of software, devoid of specialized hardware requirements), while being accurate, readily accessible, and user-friendly, to help software developers to easily measure and improve the power efficiency of their software.

\emph{Ergasheva and al. 2020} \cite{ergasheva2020}.
The authors emphasize the necessity of evaluating energy consumption at any stage of software production. With this objective in mind, they aim at building and validating a quantitative framework that can guide the development and evolution of sustainable software systems. This framework would rely on a diverse range of metrics gathered throughout the software systems' life cycle, and optimize system performance according to relevant factors such as the efficient utilization of resources. Among the reviewed metrics and tools, the following are related to our research questions: PETrA \cite{nucci2017}, Green-Advisor \cite{aggarwal2015}, ePRO-MP \cite{wonil2009}, and ANEPROF \cite{chung2011}.

\begin{table}[] 
\centering 
\input{tables/subsidiary_3}
\caption{Some tools and methods mentioned in selected secondary studies (with associated scientific study)} 
\label{tab:in_secondary_not_alone} 
\end{table}

\begin{table}[] 
\centering 
\input{tables/subsidiary_2}
\caption{Some tools and methods mentioned in selected secondary studies (without associated scientific study)}
\label{tab:in_secondary_alone} 
\end{table}

\emph{Fahad and al. 2019} \cite{fahad2019}.
In a review situated between a survey and an experimental study, the authors examine approaches that enable the estimation of dynamic energy consumption in software. They emphasize the importance of utilizing dynamic (as opposed to static) energy for optimizing application energy consumption. They concentrate on the accuracy of energy estimation techniques, considering it key for optimization, when compared to actual measurements taken by \acrshort{epm}s.
The authors distinguish between the following types of approaches: system-level physical measurements using \acrshort{epm}s, measurements using on-chip power sensors, energy predictive models. Concerning the latter, they mention that most models are linear and based on \acrshort{pmc}s. In the experimental study, dynamic energy measured by \acrshort{epm} is compared with dynamic energy consumption ``estimated'' in three different ways: by \acrshort{rapl} alone, by a combination of \acrshort{nvml} and the Intel-\acrshort{smc} equipped in Intel Xeon Phi co-processors, and by their own prediction models. The latter consist of six distinct linear regression models, utilizing commonly used \acrshort{pmc}s, based on several references reviewed by the authors. They notably conclude that relying on inaccurate energy measurements provided by on-chip sensors for dynamic energy optimization can result in significant energy losses of up to 84\%.

\emph{Rieger and al. 2017} \cite{rieger2017}.  The authors survey research on assessing the energy consumption of software systems and review directly usable tools for programmers. The authors initially focused on tools that enable the understanding of a program's energy behavior but found limited results on this topic. They examine two kinds of approaches, both referred to as energy measurement. The first group comprises approaches that are based on measurements (as defined in Section \ref{sec:taxonomy}). The second group involves approaches that derive a model of energy consumption (model parameters are still set on the basis of measurements) and then enables the estimation of a program's energy behavior (without direct measurement) before its execution on a device. For the first category, they review the following tools: Silicon Labs, SEFLab \cite{ferreira2013}, JouleUnit \cite{wilke2013}, Green-Mining \cite{hindle2012}, and Greendroid \cite{couto2015}. For the second category, they review the following tools: PowerTutor \cite{zhang2010} and five different power models: for the power consumption model of a general-purpose computer, for a program’s energy consumption at design time, for the energy used for each CPU instruction, for energy consumption analysis at the instruction level, and for mapping source code to energy consumption. They emphasize the lack of fine-grained measurement approaches as well as approaches for general-purpose platforms.


\emph{Mobius and al. 2014} \cite{mobius2014}.
The authors provide a comprehensive survey on the energy consumption of single-core and multi-core processors, virtual machines, and entire servers. Their focus is on reducing the energy consumption of Internet servers and data centers, addressing the lack of proportionality between this consumption and the work accomplished. They emphasize that the accurate estimation of a server's energy consumption and its subsystems is central to solving this issue. The authors also distinguish between energy measurement and estimation approaches (the latter termed as ``power estimation models'' in \cite{mobius2014}). After outlining energy measurement methods briefly, the authors review seven models for CPUs, six models for virtual machines, and seven models for entire servers. They conclude that most existing models utilize hardware performance counters as input and employ regression techniques. Consequently, they consider \acrshort{pmc}s and benchmarks as essential components in a wide range of power estimation models. They discuss factors influencing the estimation error, such as the choice of model input parameters, model training techniques, training data, and reference power. Furthermore, they note that current models are primarily limited to static workloads (i.e., workloads that mostly remain constant) and advocate the need to develop models for dynamic workloads. Finally, they highlight the importance of balancing the accuracy of the energy estimation model with its complexity (resource consumption) and estimation latency.

\emph{Noureddine and al. 2013} \cite{noureddine2013}. 
The authors address energy estimation for energy management, encompassing a spectrum from reducing software and hardware usage to compiler optimization, from server consolidation to software migration. They differentiate between measurement techniques (referred to as ``monitoring the energy consumption of hardware components'') and estimation techniques (referred to as ``estimating the energy consumption of hardware and software''). They term the calculation formulas upon which the latter techniques are often based as ``power models.'' Estimation is further categorized into two groups: the first employs power models, while the second, termed ``software measurement,'' relies on statistical sampling or software code instrumentation. After discussing energy modeling for hardware and software consumption, they notably review several tools for estimating software energy consumption: PowerScope \cite{flinn1999}, pTop \cite{do2009}, Jalen \cite{noureddine2014a}, and PowerAPI \cite{bourdon2013}. Finally, the authors conclude that more work is required to develop energy estimation approaches that are 1) accurate, for improved precision in energy optimization, 2) fine-grained, for clearer insights into how and where energy is being spent, 3) more software-centric, for increased flexibility, evolution, and reusability, and 4) with a smaller impact on user experience (i.e., low computational overhead, no need for additional hardware or manual modification of the application's code), to enhance the usability and adoption of energy measurement tools.

\subsubsection{Surveys Focused on ML}

\emph{Bannour and al. 2021} \cite{bannour2021}. 
The authors take a quantitative approach, reviewing available tools capable of evaluating the energy consumption and CO2 emissions of \acrshort{nlp} algorithms. They establish specific inclusion and exclusion criteria for the tools under review and testing: the tool must be freely available, usable in their programming environment (Mac/Linux terminal), documented in a scientific publication, suitable for measuring the impact of \acrshort{nlp} experiments (such as Named Entity Recognition), and capable of providing a CO2 equivalent measure for experiments. Specifically, the selected tools must account for both CPU and GPU consumption. The authors review six tools: Carbon-Tracker \cite{anthony2020}, Experiment-Impact-Tracker \cite{henderson2020}, Green-Algorithms \cite{lannelongue2021}, ML-CO2-Impact \cite{lacoste2019}, Energy-Usage \cite{lottick2019}, and Cumulator \cite{trebaol2020}. They quantitatively compare the outputs of these tools in Named Entity Recognition experiments conducted on various computational setups (local server, computing facility). The authors observe discrepancies among the outputs of these tools and provide potential causes for these differences. However, they conclude that more work is necessary to better understand these discrepancies.

\emph{Garc\'ia-Mart\'in and al. 2019} \cite{garcia-martin2019b} and \cite{garcia-martin2019a} (extension of \cite{garcia-martin2019b}).
The authors survey power estimation approaches developed in the computer architecture community that could be applied to machine learning use cases. Observing a lack of power models in existing ML frameworks such as TensorFlow or Caffe, the authors aimed to guide the machine learning community, providing them with knowledge and tools to construct their own energy models. They employed the following taxonomy to classify power models. One category encompasses ``software-level models'' (focused on the energy consumption of the application or software implementation), which further includes abstract ``application-level models'' (linking application characteristics, like the number of parameters in a neural network, to energy consumption) and ``instruction-level models'' (linking program instructions to energy consumption, where ``execution traces'' might be simulated or captured by \acrshort{pmc}s, for instance). The other category involves ``hardware-level or functional-level models'' (connecting specific hardware components to energy consumption, identifying which hardware strongly correlates with the power used by the application). Consequently, they reviewed power estimation models, notably those designed for CPUs and RAM. They assessed 23 models, 5 tools facilitating the creation of power models (ARM-Streamline \cite{rodrigues2017}, Powmon \cite{walker2017}, Intel-Power-Gadget, McPAT \cite{li2009}, PAPI \cite{weaver2012}), and finally, 5 models specifically developed for ML (including SyNERGY \cite{rodrigues2018}, Neuralpower \cite{cai2017}, and DeLight \cite{rouhani2016}). Additionally, the authors emphasized the necessity of energy models for GPUs, given that ML models are commonly trained on these platforms.


\section{Experimental Comparison of a Subset of Methods and Tools}
\label{sec:experiments}

In this section we are comparing a subset of energy consumption evaluation tools and methods on different ML computing tasks: the training or fine-tuning of ML models for computer vision and \acrshort{nlp}. 
In the different ML contexts, we observe the relative energy consumption evaluation provided by these tools and methods (also compared to an external power meter), as they belong to different approaches: on-chip sensors, mixed on-chip sensors and analytical estimation model, and two different types of analytical estimation models. While the process of constructing data-based estimation models may be available in the literature, we did not find any open-source models, and thus, they are not included in this experiment.

\begin{table}
\centering
\begin{tabular}{{|p{4cm}|p{8cm}|}}
\toprule
\bfseries Computer & Alienware Desktop Computer \\
\bfseries CPU & Intel Core i9-9900K CPU - 3.60GHz - with 8 cores (2 threads per core: 16 virtual cores) - \acrshort{tdp} 95W \\
\bfseries RAM & 64,0 GB \\
\bfseries Integrated GPU & Intel UHD Graphics 630 \\
\bfseries GPU 1 & NVIDIA GeForce RTX 2080 SUPER - \acrshort{tdp} 250W \\
\bfseries GPU 2 & NVIDIA GeForce RTX 2080 SUPER - \acrshort{tdp} 250W \\
\bfseries OS Version & Ubuntu 22.04.1 LTS (Jammy Jellyfish) \\
\bfseries Python & 3.10.6 \\
\bottomrule
\end{tabular}
\caption{Computer information}
\label{tab:computer_info}
\end{table}

\subsection{ML Computing Tasks}
The machine learning computing tasks are implemented in the PyTorch framework for the image classification tasks and in Tensorflow for the \acrshort{nlp} task. As mentioned before, the code used for these experiments is available on \href{https://github.com/Accenture/Labs-Sustainable-AI/tree/nrj_eval_comparison}{GitHub}.
Our computing environment for these experiments is a desktop computer with an Intel CPU and two Nvidia GPUs. Only one of the two available GPUs is used for training/fine-tuning. For more detail on the environment, we refer to Table \ref{tab:computer_info}.

We have chosen 4 different ML computing tasks, that we call ``MNIST,'' ``CIFAR10,'' ``ImageNet'' and ``SQUAD'', and are as follows:
\begin{itemize}
\item \emph{Training an image classifier on the MNIST dataset.} Our reference training script is the PyTorch example ``Basic MNIST Example'' (\href{https://pytorch.org/examples/}{\texttt{https://pytorch.org/examples/}}), for image classification using ConvNets, available on GitHub in the repository \href{https://github.com/pytorch/examples/tree/main/mnist}{\texttt{pytorch/examples/tree/main/mnist}}.

\item \emph{Training an image classifier on the CIFAR10 dataset.} Our reference training script is the PyTorch tutorial ``Training a classifier'', part of ``Deep Learning with PyTorch: A 60 Minute Blitz,'' available on the pytorch website at \href{https://pytorch.org/tutorials/beginner/blitz/cifar10_tutorial.html}{\texttt{tutorials/beginner/blitz/}}.

\item \emph{Training Resnet18 on the ImageNet dataset.} Our reference training script is the recipe for training Resnet18 on ImageNet, provided by PyTorch. The corresponding code is available in the repository \href{https://github.com/pytorch/vision/tree/732551701c326b8338887a3812d189c845ff28a5/references/classification}{\texttt{pytorch/vision/references/classification/}}.

\item \emph{Fine-tuning Bert-base on the SQUADv1.1 dataset.} Our reference training script is the recipe for fine-tuning Bert-base (uncased) on the dataset SQUADv1-1, provided by google-research. It is available on GitHub in the repository \href{https://github.com/google-research/bert}{\texttt{google-research/bert/}}.
\end{itemize}

\subsection{Experiments}

\begin{figure}[]
    \centering
    \includegraphics[height = 6.45cm]{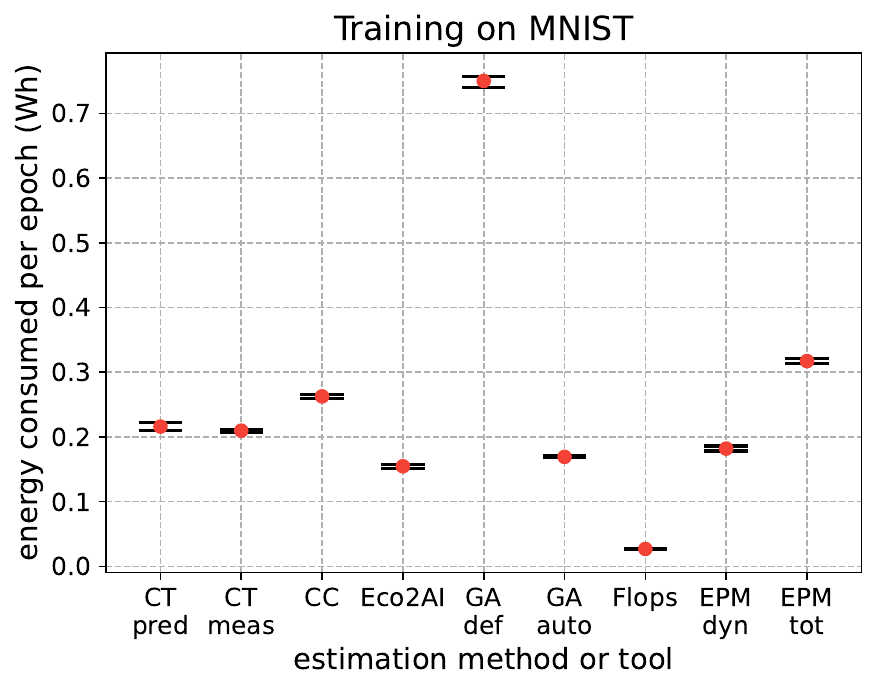}%
    \includegraphics[height = 6.45cm]{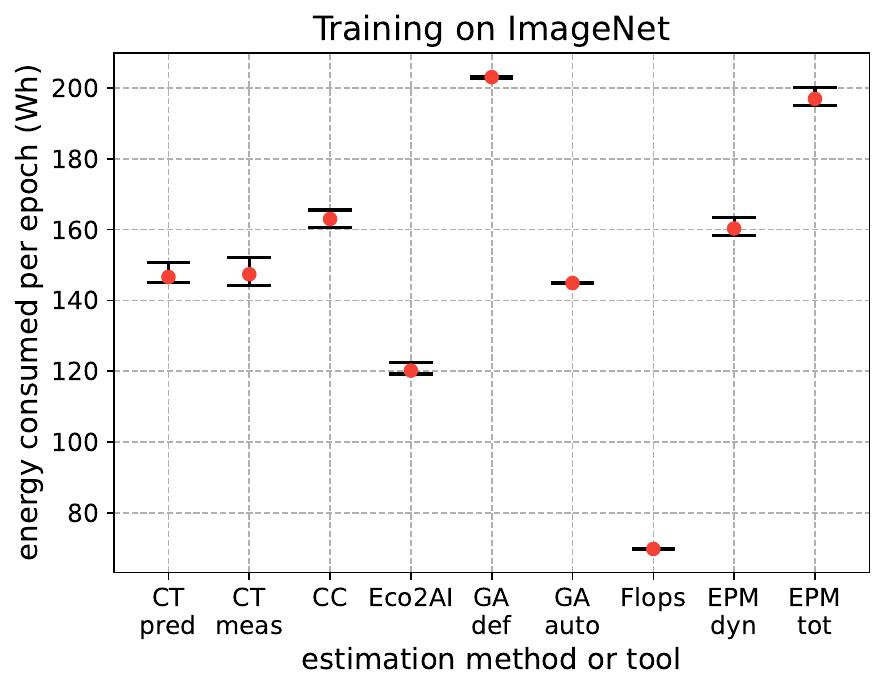}\\
    \includegraphics[height = 6.45cm]{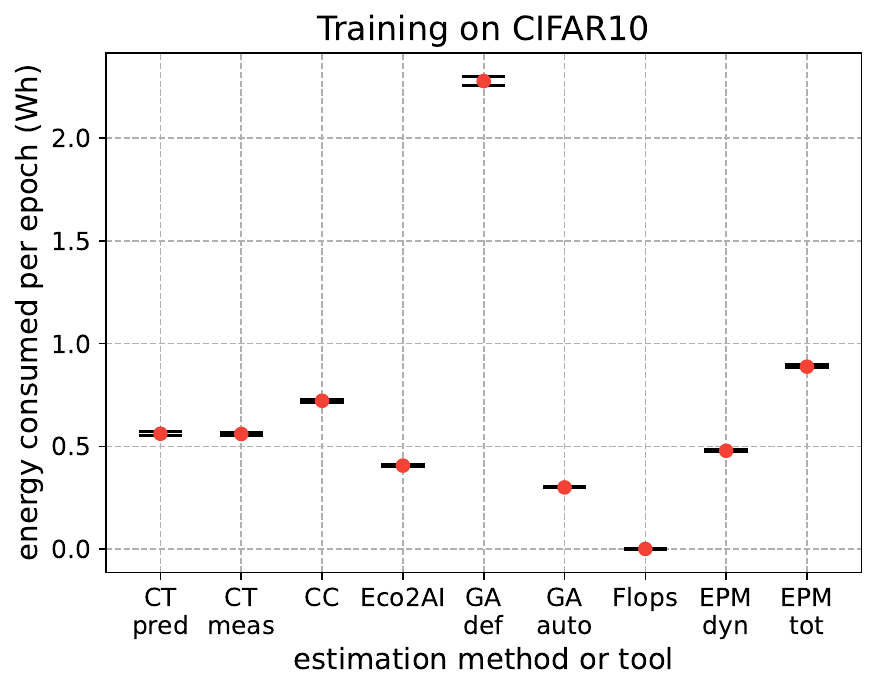}%
    \includegraphics[height = 6.45cm]{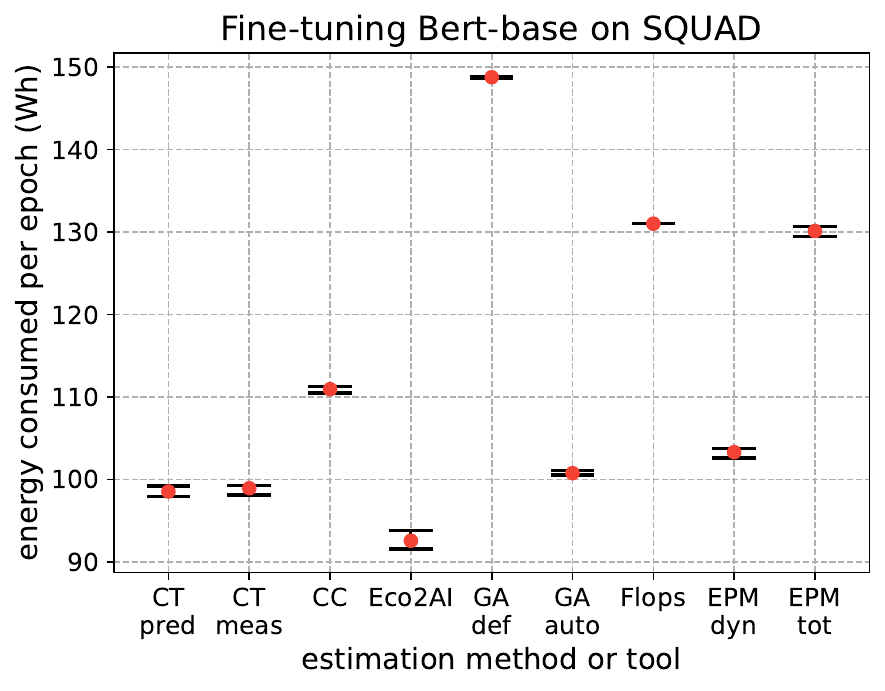}
    \caption{Energy consumed during 4 ML training tasks of different nature (vision and \acrshort{nlp}) and computational complexity (small and large models/training datasets).}
    \label{fig:res_ml}
\end{figure}

For each of the four ML computing tasks, we compare nine values of energy consumption: (1) \textbf{CT pred}: Carbon-Tracker in prediction mode (where only the first epoch of the training is monitored by the tool and the output of CT is then an estimation based on this monitoring), (2) \textbf{CT meas}: Carbon-Tracker in measurement mode (where the complete training is monitored), (3) \textbf{CC}: Code-Carbon, (4) \textbf{Eco2AI}, (5) \textbf{GA def}: Green-Algorithms with default hardware utilization rates (of CPU, RAM, and GPU), (6) \textbf{GA auto}: Green-Algorithms with monitored hardware utilization rates, (7) \textbf{Flops}: based on an estimation of the total number of \acrshort{flop}s needed for the training, the GPU throughput in \acrshort{flop}s per second, and the GPU \acrshort{tdp}, (8) \textbf{\acrshort{epm} tot}: total energy consumption evaluated by an \acrfull{epm}, (9) \textbf{\acrshort{epm} dyn}: dynamic energy consumption (i.e., the difference between the total energy consumption and the idle energy consumption, also discussed at the end of this section) evaluated by the \acrshort{epm}.

The \acrshort{epm} used here is the the smart plug ``Tapo P110'' from Tp-Link. It measures the power draw at the power outlet of the computer tower (excluding the screen). From this smart plug, we query power draw values every two seconds in parallel to the computing task. We consider \acrshort{epm} dyn as a reference value, to evaluate the precision of the other methods and tools.

We have set the other tools (CT, CC, Eco2AI, GA auto) to also query information (from the hardware or OS) every two seconds.
%
%
In the case of GA auto, we run, in parallel to the training, a python script querying the CPU, RAM, and GPU utilization (every two seconds). We then use as input to GA the mean hardware utilization across the whole training.
FLOP \cite{desislavov2023} and GA \cite{lannelongue2021} are discussed in Paragraph \ref{tab:YY-analytical-estimation-model}, while CT \cite{anthony2020} and CC \cite{lottick2019} are discussed in Paragraph \ref{tab:YY-on-chip-sensors}, and Eco2AI \cite{budennyy2022} in Paragraph \ref{tab:YY-analytical-estimation-model-and-on-chip-sensors}.

For each ML computing task and each evaluation method among CT pred, CT meas, CC, ECO2AI, GA def, GA auto, \acrshort{epm} dyn, and \acrshort{epm} tot, we train the corresponding model five times. The order of the experiments has been randomized. We record the duration of the ML computing task and energy consumed evaluation (output of the method or tool). Note that the Flops evaluation method does not require running the computing task.

\begin{figure}[]
    \centering
    \includegraphics[height = 6.45cm]{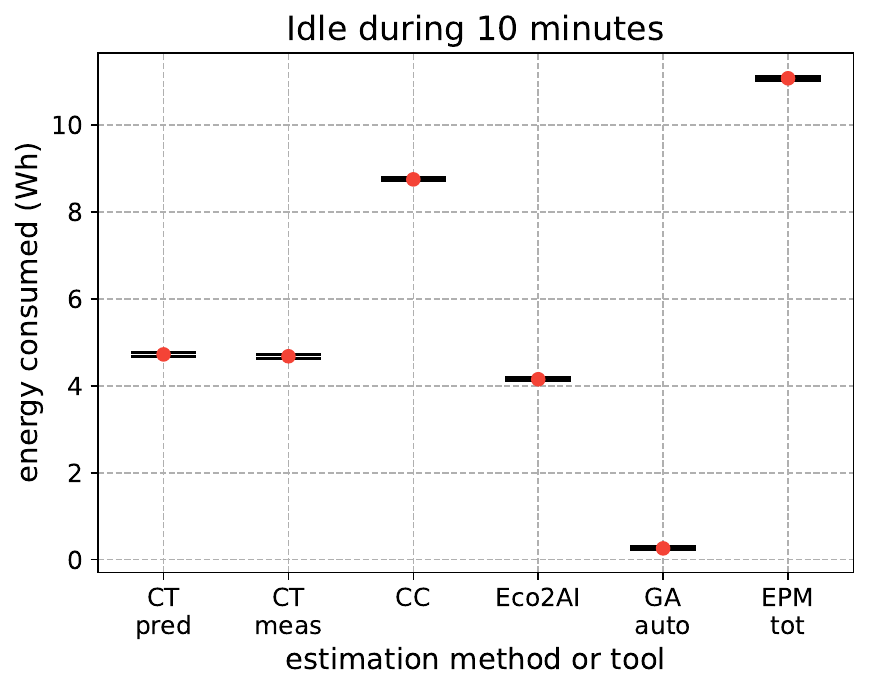}
    \caption{Idle energy consumed during 10 minutes}
    \label{fig:res_idle}
\end{figure}

We present in Figure \ref{fig:res_ml} the energy consumed in Watt-hour (Wh) per epoch as estimated by each method or tool for all four ML computing tasks. The mean across all five iterations is represented by the dot, while the minimum and maximum values across the five iteration are represented by the horizontal bars.
We display in Figure \ref{fig:res_idle} the energy consumed during 10 minutes where no application is running on the computer and the computer is not in sleep mode, which we call here ``idle energy consumed.''
This energy needs only to be evaluated for six of the tools and methods: CT pred, CT pred, CC, Eco2AI, GA auto and \acrshort{epm} tot. Indeed, for the Flop method the idle consumption is equal to zero by definition as no \acrshort{flop} are being executed, while GA def cannot capture idle consumption as it uses default hardware utilization rates.
The values presented in Figure \ref{fig:res_idle} correspond to power draws of 28 W for CT pred, 28 W for CT meas, 52 W for CC on, 25 W for Eco2AI, 2 W for GA auto, 66 W for \acrshort{epm} tot.

\subsection{Observations}


\emph{On individual methods.} We recall that CT and CC are both based on \acrshort{rapl} and \acrshort{nvml}. However, CT provides process-level estimation, unlike CC, and should thus better identify the consumption due to training. This may explain the smaller output value for CT. Eco2AI, also a process-level tool, uses an analytical estimation model based on \acrshort{tdp} and hardware utilization for the CPU and RAM, as well as \acrshort{nvml} for the GPU.

GA, however, is solely based on an analytical estimation model that accounts for the CPU, RAM, and GPU, depending on \acrshort{tdp} and hardware utilization. The higher value for GA def is due to the default hardware utilization of GA being set at 100\%. The relative gap between GA def and GA auto (and other evaluation tools and methods) becomes less pronounced when compared on the training of ResNet18 or the fine-tuning of Bert-base, which are more computationally demanding.

The Flops method behaves very differently for computer vision tasks compared to the \acrshort{nlp} task. Indeed, for the former, the consumption output of the Flops method is significantly smaller than the estimates from all other evaluation methods and tools. This may be explained by the way energy is consumed during training in computer vision versus \acrshort{nlp}. In computer vision, the data type and the use of convolutional layers may involve a lot of data movement not accounted for by Flops. In contrast, in \acrshort{nlp}, the machine learning model itself is large, which may lead to high power consumption for computation. However, this interpretation should be approached with caution, as the Flops method assumes that all operations are executed on the GPU to estimate consumption during inference and relies on an approximation to extrapolate the inference consumption to training.

The \acrshort{epm} tot energy consumption output is not directly comparable with evaluation methods that estimate CPU, RAM, and GPU power draws, as \acrshort{epm} measures the energy consumption of the entire computer, including, for example, the one due to fans. Indeed, \acrshort{epm} total values are higher than those from all other evaluation methods, except for GA def, which assumes that the system operates at maximum intensity.

\emph{On the relative order between methods.}
All methods are relatively stable across iterations. The relative order of the evaluation methods is generally preserved across computing tasks, except for GA auto and Eco2AI. In these cases, the output of GA auto may be higher or lower than that of Eco2AI, depending on the task.

CT meas, CC, and GA auto produce progressively larger energy estimates, consistent with the observations in \cite{jay2023} (these are the common tools across the two surveys). The only exception is for fine-tuning Bert-base, where the order between the evaluation tools CT meas, CC, and GA auto changes. The same observation holds for CT and GA auto\footnote{We believe that the authors in \cite{jay2023, bannour2021} did not use default hardware utilization inputs, though they may have used a different method than ours to approximate these inputs.} in the experiments of \cite{bannour2021}.

\emph{On the comparison of methods to the reference.}
When compared to our reference \acrshort{epm} dyn, we observe that CT (pred/meas), CC and GA auto are consistently closer to the reference than the other methods. Eco2AI is also close to the reference, but seems to perform worse on the ImageNet and SQUAD tasks. GA auto, which does not rely on on-chip sensors but has accurate information about hardware utilization, seems to perform as well as CT and CC, especially for the ImageNet and SQUAD tasks. On the contrary, GA def and Flops are consistently far from the reference.



\section{Conclusion and Outlook}
\label{sec:conclusion}

In this article, we have conducted a \acrlong{slr} of all available tools and methods to evaluate the energy consumption of machine learning. As energy consumption in computing has been a concern for a long time outside of the ML community, we included tools and methods not necessarily designed for ML, but for software in general. Our search process thus used keywords ranging from ``software'' and ``virtual machine,'' to ``deep learning'' and ``\acrshort{nlp}''. Consequently, the search led to a large number of results, requiring the development of scripts to assist in gathering and post-processing the results (available on \href{https://github.com/Accenture/Labs-Sustainable-AI/tree/slr_tools}{GitHub}). We selected a total of 118 results (primary and secondary studies) in the review process, and organized them according to a simple taxonomy: tools based on 1) measurements of actual power, current intensity, or voltage, 2) estimation models that relate indirect evidences such as activity factors or characteristics of the software to energy consumption, and that may be either data-based or analytical models, and 3) on-chip sensors. Additionally, we conducted experiments involving five tools and methods, applied on ML computing tasks (training, fine-tuning) of different nature (vision, \acrshort{nlp}) and complexity (small or large datasets and models). We observed, in the different ML contexts, the relative energy consumption evaluation provided by these tools and methods, also compared to an external power meter. These experiments can be done on a desktop computer equipped with a single GPU, and the scripts are available on \href{https://github.com/Accenture/Labs-Sustainable-AI/tree/nrj_eval_comparison}{GitHub}.

We highlight some limitations of our work. Firstly, as mentioned above, the large scope of our search led to a high number of results with a single search by means of keywords. Due to this, we did not realize backward and forward searches (i.e., searching within the references and citations of the results, respectively), and no mitigation was implemented. Despite this limitation, we believe that this initial search provides by itself a broad overview of the field, and definitely broader than previous surveys (see Figure \ref{fig:heatmap}). Moreover, we remark that the availability of our scripts and data (on \href{https://github.com/Accenture/Labs-Sustainable-AI/tree/slr_tools}{GitHub}) make it possible to extend our analysis.
Concerning the experiments, interesting extensions include experiments with more varied ML training, as well as inference tasks, and with data-based estimation models, that have not been included due to the lack of available open-source models.

We would also like to comment on some interesting features and limitations of the reviewed tools and methods.
Overall, we have observed that tools based on on-chip sensors are considered sufficiently precise for many use cases, and several studies use them as training target for data-based estimation models (e.g., \cite{morlans2022, alavani2023}). Our experiments seem to confirm that such methods provide results close to the reference.
In \cite{getzner2023}, the authors use Code-Carbon itself as the target, though they warn that this may introduce errors because Code-Carbon could lack precision for their use case (as it does not isolate processes).
However, \cite{fahad2019} comments that the accuracy of on-chip sensors is not sufficient for dynamic energy optimization use cases, and several studies mention the lack of clarity on the underlying technique and accuracy of on-chip sensors.
Several tools have been developed based on on-chip sensors in recent years, often with applications to machine learning in mind. They were at first developed for Linux OS only, but this is changing, and tools such as Code Carbon are now available for various OS.
On-chip sensors-only tools may present issues in terms of supported hardware (e.g., \acrshort{rapl} and \acrshort{nvml} apply to Intel CPUs and Nvidia GPUs only, respectively), or the need for administrative rights on the monitored machine (e.g., to access \acrshort{rapl} data). However, solutions or alternatives are being developed. For instance, the recently developed tool Eco2AI uses an analytical estimation model for the CPU consumption and thus bypasses the aforementioned problems related to the CPU. Nevertheless, our experiments show that the output of Eco2AI is systematically far from on-chip-sensor approaches and the \acrshort{epm} for heavy computational tasks, and further study is needed to understand this result. Moreover, new tools are being developed to support more hardware. This is the case with Energy-Scopium (Table \ref{tab:in_secondary_alone}), which includes AMD CPUs, or PMT \cite{corda2022} which supports AMD GPUs. Tools based on analytical estimation models such as GA \cite{lannelongue2021} also bypass hardware compatibility issues for the GPU.
In addition, some of these tools (both from on-chip sensors and analytical estimation model approaches) can isolate a specific function or process of a given script. This is the case, for example, with Carbon-Tracker \cite{anthony2020} and Eco2AI \cite{budennyy2022}.
An open question is whether the evaluations provided by these readily available tools are relevant in a cloud environment (where a single CPU is shared by several virtual machines), as there is a lack of clarity on what data they actually access (when recovering CPU utilization or \acrshort{rapl} data, for instance). Green-Algorithms \cite{lannelongue2021}, which is an analytical estimation model based on hardware utilization and \acrshort{tdp}, proposes entering the \acrshort{tdp} per core of the hardware and the number of CPU cores, giving the user the possibility to declare the number of CPU cores (or ``virtual CPUs'') allocated to their instance. Besides, Kupler (see Appendix \ref{ap:subsitiary_tools_alone_not_in_selection}) and \cite{pathania2023} are working on solutions for virtual environments.

In this work, we have discussed on-chip sensors and analytical estimation model approaches, which are often relatively more accessible than data-based estimation models. Tools or methods based on data-based estimation models are very varied. Many are linear models based on \acrshort{pmc}s. However we have also seen, for instance, models for the GPU that are based on PTX code (this is code used for the compilation phase) as well as on a more abstract level, models based on characteristics of the application (e.g., a neural network's architecture), that do not require the end user to run the target computing task (once the estimation model has been developed).
Data-based estimation models are generally not publicly available, contrary to the tools mentioned earlier, and the methodology to build them is more involved. 
In \cite{shahid2021a}, the authors aim to provide guidance on how to construct linear estimation models based on PMCs. This raises questions, such as which models can be shared between computers and what constraints (e.g., computer architecture) apply. Additionally, the reported accuracy of these models may not be tested uniformly, making cross-study comparisons challenging.

Finally, as we are concerned with the energy, and more generally the environmental impact of AI, let us relay that several studies, such as \cite{bannour2021, anthony2020, henderson2020}, stress that the production and end of life of the hardware used for the target computing task is also of importance, though often left out of the scope of the concerned studies. 
Several studies \cite{fu2018, noureddine2014, noureddine2022} have highlighted the need to understand how energy is consumed in the hardware or which parts of a given code consume the most energy, and have contributed in this direction. For instance, the authors of \cite{fu2018} analyze the features of their energy estimation model in order to understand where energy efficiency may be improved. Tools that have also been developed in this direction include JalenUnit \cite{noureddine2014} and JoularJX \cite{noureddine2022}.
The process of creating new tools and methods is still ongoing, and some recent papers make recommendations on the kind of tools that would benefit the community \cite{pathania2023}.

To conclude, we share the opinion expressed by the authors of \cite{garcia-martin2019a}, who ``believe that the reasons why the machine learning community has not shown more interest in energy consumption is because of their lack of familiarity with the current approaches to estimate energy and the lack of power models in existing machine learning frameworks.'' Several other studies have taken similar positions and have been motivated to address these issues (e.g., by building accessible tools, as in \cite{lannelongue2021} and \cite{anthony2020}). We hope that this paper contributes to this effort.

\clearpage

\phantomsection
\addcontentsline{toc}{section}{Acronyms}
\printglossary[type=\acronymtype]

\clearpage

\phantomsection
\addcontentsline{toc}{section}{References}
\bibliographystyle{plain}
\bibliography{bibliography}

\begin{appendices}

\section{Appendix}

\subsection{Additional Tools not Documented within a Scientific Article}
\label{ap:subsitiary_tools_alone_not_in_selection}

\input{tables/subsidiary_1}

\subsection{Search Queries}
\label{ap:queries}

The query used for the ACM data source is as follows:
("machine learning" OR "deep learning" OR computing OR "information and communications technology" OR \acrshort{ict} OR "artificial intelligence" OR AI OR "natural language processing" OR \acrshort{nlp} OR "neural network" OR "neural networks" OR \acrshort{cnn} OR DNN OR computation OR computations OR software OR "process-level" OR server OR "virtual machine" OR "federated learning" OR "distributed learning") AND (measure OR measuring OR estimate OR estimation OR consumed OR consumption OR predict OR prediction OR predicting OR track OR tracking OR report OR reports OR reporting OR account OR quantify OR quantifying OR monitor OR monitoring OR evaluate OR evaluating) AND (energy OR power OR "environmental impact" OR "carbon footprint" OR "carbon emissions" OR "carbon impact") NOT (wind OR building OR buildings OR vehicles OR homes OR ships OR solar OR photovoltaic OR vehicle).

The query used for the IEEE data source is as follows:
("Document Title":"machine learning" OR "Document Title":"deep learning" OR "Document Title":computing OR "Document Title":"information and communications technology" OR "Document Title":\acrshort{ict} OR "Document Title":"artificial intelligence" OR "Document Title":AI OR "Document Title":"natural language processing" OR "Document Title":\acrshort{nlp} OR "Document Title":"neural network" OR "Document Title":"neural networks" OR "Document Title":\acrshort{cnn} OR "Document Title":DNN OR "Document Title":computation OR "Document Title":computations OR "Document Title":software OR "Document Title":"process-level" OR "Document Title":server OR "Document Title":"virtual machine" OR "Document Title":"federated learning" OR "Document Title":"distributed learning") AND ("Document Title":measure OR "Document Title":measuring OR "Document Title":estimate OR "Document Title":estimation OR "Document Title":consumed OR "Document Title":consumption OR "Document Title":predict OR "Document Title":prediction OR "Document Title":predicting OR "Document Title":track OR "Document Title":tracking OR "Document Title":report OR "Document Title":reports OR "Document Title":reporting OR "Document Title":account OR "Document Title":quantify OR "Document Title":quantifying OR "Document Title":monitor OR "Document Title":monitoring OR "Document Title":evaluate OR "Document Title":evaluating) AND ("Document Title":energy OR "Document Title":power OR "Document Title":"environmental impact" OR "Document Title":"carbon footprint" OR "Document Title":"carbon emissions" OR "Document Title":"carbon impact") NOT ("Document Title":wind OR "Document Title":building OR "Document Title":buildings OR "Document Title":vehicles OR "Document Title":homes OR "Document Title":ships OR "Document Title":solar OR "Document Title":photovoltaic OR "Document Title":vehicle).

Finally, here is an example of one of the sub-queries used for the data source Google Scholar:  
allintitle:("machine learning") + (measure|measuring|estimate|estimation) + (energy|power|"environmental impact"|"carbon footprint"|"carbon emissions"|"carbon impact") -wind -building -buildings -vehicles -homes -ships -solar -photovoltaic -vehicle.


\subsection{List of excluding words}
\label{ap:excluding_words}

The following words and pairs of words are identified, in the semi-automated section phase, as ``excluding words'' (more detail can be found on \href{https://github.com/Accenture/Labs-Sustainable-AI/tree/slr_tools}{GitHub}):
absorptiometry, absorption, accident, accidents, acids, acoustic, adenocarcinoma, adolescents, adults, aerial, africa, african, agricultural, agriculture, air, aircraft, alloy, alloys, amino, amsterdam, animal, anomaly detection, ant, applications, aquifers, arabia, arena, art, asian, atlas, atmospheric, atom, atomic, atomistic, atomization, atoms, autoimmune, automotive, bankruptcy, batteries, battery, batteryless, beijing, bimetallic, biodiesel, biofuels, biogas, biological, biomass, biomedical, bipv, boiler, boilers, bone, brain, brazilian, broiler, broilers, business, businesses, calorimeter, campus, canadian, cancer, car, carbonyls, carcass, carcinoma, cardiac, cardiomyopathy, cardiopulmonary, cardiovascular, cargo, carrier, cars, catalytic, cells, cement, centrifugal, cervical, chamber, chambers, charcoal, chemical, chemistry, children, china, chinese, chips, chromatography, city, cleanroom, climatic, clinical, coal, coastal, cognitive, combustion, communities, commuter, companies, condenser, condition monitoring, congress, converter, converters, conveyor, copper, corn, corneal, covid, crop, cryptographic, crystals, cucumber, cyclotron, dam, daylighting, deforestation, desulfurization, diagnose, diagnosing, diagnosis, diagnostic, diagnostics, diesel, dietary, digestible, disaster, disease, diseases, distribution line, domestic, dosimetric, driver, drone, drones, drug, drugs, dryer, drying, dual energy, dust, ecological, economic, economics, economies, ecosystems, electrochemical, electrolysis, electromagnetic, electron, electrons, emergency, energy expanditurefree energy estimation, energy expenditure, energy generation, energy harvesting, energy production, energy storage, energy system, energy systems, enterprise, enterprises, enthalpy, enzyme, epileptic, erosion, evapotranspiration, expenditure estimation, facilities, factory, failure, failures, farms, fault detection, fault prediction, fibrillation, financial, fire, firefly, flame, fleet, flight, flood, flue, fluid, fluidized, fluids, flying, fog, food, foods, forest, forests, fracture, fraud detection, freshwater, frozen, fuel, fuels, fusion, garbage, gas, gases, gasification, gasoline, gastric, genes, genetically, geomechanical, geostationary, geosynchronous, geothermal, german, germany, glucose, glycoprotein, graphene, gravitational, greek, greenhouse, greenhouses, grid, grids, grinding, groundwater, harnessing, harvester, harvesting, health, healthcare, heart, heat, heating, heatwave, heatwaves, hip, home, homeostasis, homeostatic, hospital, hospitalised, hospitals, hotel, house, houseec, household, householders, households, houses, housing, human, humidity, hybrid energy, hydraulic, hydro, hydrocarbon, hydrodynamic, hydrodynamics, hydroelectric, hydroelectricity, hydrogen, hydrolase, hydrologic, hydropower, hydrostatic, hydrothermal, hydroturbine, hyperspectral, ice, immune, india, indian, indonesia, indoor, indoors, industrial, industrialization, industries, industry, injection, insulators, internuclear, intramolecular, intraocular, ion, ionic, ionization, iran, iranian, island, japan, june, kinetic, kinetics, korea, laboratory, lamb, land, languages, lanka, laser, leakage, ligand, liquefaction, lithium, lung, lyapunov, lymph, machinery, macroeconomic, magnet, magnetic, magnetics, malaysia, malware, manufacturing, maritime, market, marketers, marketing, markets, mars, materials, matter, meal, measuringconverters, meat, mechanical, medical, membrane, membranes, metabolizable, metaheuristic, metal forming processes, metallurgical, metals, metastasis, meteorological, methionine, metropolitan, microalgae, microgrid, microgrids, microwave, milk, mill, milling, mills, mine, mining, mmwave, moisture, molecular, molecule, molecules, moroccan, morocco, motor, motors, mountain, muav, multiphysics, multispectral, mushroom, myocardial, myopia, nanoflakes, nanofluid, nanofluids, nanoparticles, netherlands, neuroimmune, neuronal, neutron, neutronic, neutronics, neutrons, nitrogen, nuclear, ocean, office, ofhousehold, oil, ontology, orbit, orbiting, outage, outages, outdoor, overvoltages, oxygen, oxygenate, pandemic, paper, particle, particles, permeation, petrochemical, petroleum, photonic, photoplethysmography, photovoltaics, pilot, plane, plant, plants, plasma, plasticity, plethysmography, pollutant, pollutants, pollution, polymer, polystyrene, population, portuguese, postal, potato, poultry, poverty, power distribution network, power distribution system, power flow, power generation, power line, power load, power loss, power quality, power spectrum, power system, power systems, power transformer, powerplants, pressure, price prediction, prognosis, prognostic, property, propulsion, protein, proteins, province, provinces, psychological, pump, pumped, pumping, pumps, pv, pvt, pvusa, pyrolysis, qatar, quality monitoring, radiation, radiative, railway, railways, rainfall, reactive power, reactor, reactors, refactoring, renewable energy, reservoir, reservoirs, residences, residential, resources, respiratory, risk, risks, river, robot, robots, romania, room, rooms, rotor, rural, russia, satellite, satellites, saudi, scanner, school, scooter, screw, sea, seawater, seismic, seizure, sequencing, shear walls, ship, shipboard, shower, showers, skating, skiers, sky, smart cities, socioeconomic, solid, solvents, space, spaceborne, spacecraft, spatiotemporal, species, spectra, spectral, spectroscopy, spintronics, spv, sri, stability monitoring, stability prediction, state estimation, steel, steelmaking, steels, stereolithography, storm, suburban, sugar, sugarcane, sulfur, superalloy, superalloys, supergrids, supermarket, swimming, taiwan, tehran, telescope, therapy, thermal, thermochemical, thermoelectric, thermomechanical, thermoplastic, tidal, tomography, tourism, tractor, tractors, traffic, trains, transient stability, transmission line, transmission system, trucks, turbine, turbomachinery, turkey, tyre, uav, uavs, ultrasonic, ultrasound, underwater, urban, urbanization, us, vacuum, vehicular, vessel, vessels, vibration, vibrations, voltage stability, voltaic, warehouses, warning, waste, wastes, wastewater, water, wave, waveform, waveforms, wavelength, wavelet, wavelets, wavenet, waves, weather, westinghouse, wheat, wheelchair, wildfire, wildlife, wood, woodland, woodworking, zealand.


\subsection{Full URLs}
\label{ap:links}

Links for primary studies description ($\mathscr{{Y}Y}$, $\mathscr{{Y}N}$):

\input{tables/linklist}

Links for tools not documented in a scientific article and not cited in the selected secondary studies:

\input{tables/subsidiary_linklist_df1}

Links for other tools not documented in a scientific article:

\input{tables/subsidiary_linklist_df2}

\end{appendices}
\end{document}

%% file: glossary/glossary.tex
\newacronym{ict}{ICT}{Information and Communication Technology}
\newacronym{nlp}{NLP}{Natural Language Processing}
\newacronym{nas}{NAS}{Neural Architecture Search}
\newacronym{slr}{SLR}{Systematic Literature Review}
\newacronym{tdp}{TDP}{Thermal Design Power}
\newacronym{epm}{EPM}{External Power Meter}
\newacronym{mac}{MAC}{Multiply-Accumulate}
\newacronym{flop}{FLOP}{Floating Point Operation}
\newacronym{pmc}{PMC}{Performance Monitoring Counter}
\newacronym{nn}{NN}{Neural Network}
\newacronym{rapl}{RAPL}{Running Average Power Limit}
\newacronym{nvml}{NVML}{Nvidia Management Library}
\newacronym{smc}{SMC}{Intel System Management Controller chip}
\newacronym{ptx}{PTX}{Parallel Thread Execution}
\newacronym{cnn}{CNN}{Convolutional Neural Network}
\newacronym{nu}{NU}{Name Unspecified}

%% file: tables/YY-timeline.tex
\begin{tikzpicture} 
\draw[->, thick] (0,0) -- (15,0); 
\foreach \x/\year in {0/2017, 2/2018, 4/2019, 6/2020, 8/2021, 10/2022, 12/2023} { 
    \draw[shift={(\x,0)}, color=black] (0pt,2pt) -- (0pt,-2pt); 
    \node[below] at (\x, 0) {\year}; 
    } 
\node[above right, align=left, rotate=45] at (0,0) {DNNEET \cite{yang2017}\\ NeuralP \cite{cai2017}}; 
\node[above right, align=left, rotate=45] at (2,0) {SyNERGY \cite{rodrigues2018}}; 
\node[above right, align=left, rotate=45] at (4,0) {CC \cite{lottick2019}\\ MLCI \cite{lacoste2019}}; 
\node[above right, align=left, rotate=45] at (6,0) {CT \cite{anthony2020}, EIT \cite{henderson2020}\\ Cumulator \cite{trebaol2020}}; 
\node[above right, align=left, rotate=45] at (8.3,0) {Accelergy \cite{wang2021}, EcoML \cite{igescu2021}\\ EnergyNN \cite{goel2021}, GA \cite{lannelongue2021}\\ NU \cite{metz2021}, Pommel \cite{montgomerie-corcoran2021}}; 
\node[above right, align=left, rotate=45] at (10,0) {BT \cite{carastan-santos2022}, Eco2AI \cite{budennyy2022}\\ NU \cite{lemaire2022} \cite{metz2022b} \cite{metz2022a} \cite{lahmer2022}}; 
\node[above right, align=left, rotate=45] at (12,0) {NU \cite{desislavov2023} \cite{qiu2023} \cite{dariol2023} \cite{ortega2023}\\ DLEE \cite{getzner2023}, ZeusM \cite{you2023}}; 
\end{tikzpicture}

%% file: tables/YY-description.tex
\ref{tab:YY-analytical-estimation-model} for the group `analytical estimation model', \ref{tab:YY-data-based-estimation-model} for the group `data-based estimation model', \ref{tab:YY-on-chip-sensors} for the group `on-chip sensors', \ref{tab:YY-analytical-and-data-based-estimation-model} for the group `analytical and data-based estimation model', \ref{tab:YY-analytical-estimation-model-and-on-chip-sensors} for the group `analytical estimation model and on-chip sensors', \ref{tab:YY-other} for studies without group.%

%% file: tables/YY-analytical-estimation-model.tex
\paragraph{YY studies in the group ``analytical estimation model''} 
 \label{tab:YY-analytical-estimation-model} 
\begin{longtable}{|>{\raggedright\arraybackslash}p{2cm}|p{5.9cm}|>{\raggedright\arraybackslash}p{1.85cm}|>{\raggedright\arraybackslash}p{1.95cm}|>{\raggedright\arraybackslash}p{1.5cm}|p{0.75cm}|}
\toprule
\bfseries study & \bfseries detail & \bfseries target task & \bfseries constraints & \bfseries available & \bfseries cites \\
\midrule 
\endhead
\cite{desislavov2023}, 2023, \acrshort{nu} & Model with \acrshort{flop} count as input, accounting for computations only, on CPU or GPU & any & no & no & 15 \\
\cite{lemaire2022}, 2022, \acrshort{nu} & Model with \acrshort{nn} architecture as inputs, uses energy consumption of single operations drawn from the literature, and memory size, accounting for CPU or accelerator & Spiking and non-spiking \acrshort{nn} inference & no & code and models upon request & 11 \\
\cite{lannelongue2021}, 2021, GA & (Green-Algorithms) Model with task duration and hardware utilization as inputs, accounting for CPU, RAM and GPU & any & no & code, API \href{https://github.com/GreenAlgorithms/green-algorithms-tool}{\ref*{link-lannelongue2021}} & 165 \\
\cite{trebaol2020}, 2020, Cumulator & Model with computing task duration as input, accounting for CPU, RAM and GPU & any & Python & PyPI package, code \href{https://github.com/epfl-iglobalhealth/cumulator}{\ref*{link-trebaol2020}} & 5 \\
\cite{lacoste2019}, 2019, MLCI & (ML-Co2-Impact) Model with task duration as input, accounting for one GPU & any & no & code, API \href{https://github.com/mlco2/impact}{\ref*{link-lacoste2019}} & 442 \\
\cite{yang2017}, 2017, D\acrshort{nn}EET, for normalized energy consumed & (Deep-Neural-Network-Energy-Estimation-Tool) Model with shape of layers, and number of non-zero values and and bitwidths in filters and feature maps as inputs (uses pre-computed dataflows to calculate the number of bits accessed at each memory level), accounting for CPU or Deep \acrshort{nn} processor (= accelerator) and RAM & \acrshort{cnn} inference & no & API \href{https://energyestimation.mit.edu/}{\ref*{link-yang2017}} & 195 \\
\bottomrule
\end{longtable}

%% file: tables/YY-data-based-estimation-model.tex
\paragraph{YY studies in the group ``data-based estimation model''} 
 \label{tab:YY-data-based-estimation-model} 
\begin{longtable}{|>{\raggedright\arraybackslash}p{2cm}|p{5.9cm}|>{\raggedright\arraybackslash}p{1.85cm}|>{\raggedright\arraybackslash}p{1.95cm}|>{\raggedright\arraybackslash}p{1.5cm}|p{0.75cm}|}
\toprule
\bfseries study & \bfseries detail & \bfseries target task & \bfseries constraints & \bfseries available & \bfseries cites \\
\midrule 
\endhead
\cite{dariol2023}, 2023, \acrshort{nu} & Based on SystemC simulation (from the authors), regression model with simulated execution traces as inputs, accounting for CPU and RAM on FPGA/embeded devices & \acrshort{nn}s deployment & no & no & 0 \\
\cite{metz2022b}, 2022, \acrshort{nu} & Random Forest Tree regression model with \acrshort{ptx} code and \acrshort{cnn} characteristics as inputs, accounting for GPU and RAM & \acrshort{cnn} inference & cuda-based \acrshort{cnn}, Nvidia GPU & no & 3 \\
\cite{metz2022a}, 2022, \acrshort{nu} & K-Nearest Neighbor regression model with \acrshort{ptx} code and \acrshort{cnn} characteristics as inputs, accounting for GPU and RAM & \acrshort{cnn} inference & cuda-based \acrshort{cnn}, Nvidia GPU & no & 4 \\
\cite{goel2021}, 2021, Energy\acrshort{nn} & Linear regression model with \acrshort{mac} count and memory needed as inputs, accounting for Deep Learning Processor Unit (type of \acrshort{cnn} accelerator) on embedded platforms & \acrshort{cnn} training and inference & no & no & 2 \\
\cite{igescu2021}, 2021, EcoML, based on Cumulator & ML models (Decision Tree, Linear Regression, \acrshort{nn}, Random Forest) with training dataset of a given ML model as input, accounting for CPU, RAM and GPU & training for fixed set of ML models & Python, Sklearn & PyPI package, code \href{https://github.com/epfl-iglobalhealth/CS433-2021-ecoML}{\ref*{link-igescu2021}} & 0 \\
\cite{metz2021}, 2021, \acrshort{nu} & ML model (\acrshort{nn}) with GPGPU architecture and \acrshort{ptx} code as inputs, accounting for GPU & \acrshort{cnn} inference & cuda-based \acrshort{cnn}, Nvidia GPU & no & 4 \\
\cite{rodrigues2018}, 2018, SyNERGY & ML model (Linear Regression) with \acrshort{mac} count as input, accounting for CPU, RAM and peripherals of Jetson TX1 board & \acrshort{cnn} inference & Jetson TX1 board & (part of) code \href{https://github.com/Crefeda/SyNERGY }{\ref*{link-rodrigues2018}} & 41 \\
\cite{cai2017}, 2017, NeuralP & (Neuralpower) Sparse polynomial regression model with \acrshort{cnn} architecture and target platform as inputs, accounting for GPU & \acrshort{cnn} inference & no & code, model \href{https://github.com/enyac-group/NeuralPower }{\ref*{link-cai2017}} & 146 \\
\bottomrule
\end{longtable}

%% file: tables/YY-on-chip-sensors.tex
\paragraph{YY studies in the group ``on-chip sensors''} 
 \label{tab:YY-on-chip-sensors} 
\begin{longtable}{|>{\raggedright\arraybackslash}p{2cm}|p{5.9cm}|>{\raggedright\arraybackslash}p{1.85cm}|>{\raggedright\arraybackslash}p{1.95cm}|>{\raggedright\arraybackslash}p{1.5cm}|p{0.75cm}|}
\toprule
\bfseries study & \bfseries detail & \bfseries target task & \bfseries constraints & \bfseries available & \bfseries cites \\
\midrule 
\endhead
\cite{you2023}, 2023, ZeusM & (Zeus-Monitor) \acrshort{nvml}; is part of the Zeus framework, accounting for GPU & any & Python, Nvidia GPU & PyPI package, code \href{https://github.com/SymbioticLab/Zeus }{\ref*{link-you2023}} & 18 \\
\cite{carastan-santos2022}, 2022, BT, based on Cumulator & (Benchmark-Tracker) Based on Experiment-Impact-Tracker and AI Benchmark Alpha, accounting for process-level, for CPU, RAM and GPU & training, inference tasks from AI Benchmark Alpha\footnote{AI Benchmark Alpha is an open source library for evaluating AI performance of various hardware platforms, it contains training and inference scripts for various ML models.} & Linux OS, Intel CPU, Nvidia GPU, Python & PyPi package, code \href{https://github.com/phamthi1812/Benchmark-Tracker}{\ref*{link-carastan-santos2022}} & 1 \\
\cite{anthony2020}, 2020, CT & (Carbon-Tracker) \acrshort{rapl}, \acrshort{nvml}, accounting for CPU, RAM and GPU & ML training & Linux OS, Intel CPU, Nvidia GPU, Python & PyPI package, code \href{https://github.com/lfwa/carbontracker }{\ref*{link-anthony2020}} & 254 \\
\cite{henderson2020}, 2020, EIT & (Experiment-Impact-Tracker) \acrshort{rapl}, \acrshort{nvml}, accounting for process-level, for CPU, RAM and GPU & any & Linux OS, Intel CPU, Nvidia GPU, Python & PyPI package, code \href{https://github.com/Breakend/experiment-impact-tracker}{\ref*{link-henderson2020}} & 324 \\
\cite{lottick2019}, 2019, CC, previously Energy-Usage & (Code-Carbon) \acrshort{rapl}, \acrshort{nvml}, accounting for CPU, RAM and GPU & any & Linux OS, Intel CPU, Nvidia GPU, Python & PyPI packages, codes \href{https://github.com/responsibleproblemsolving/energy-usage}{\ref*{link-lottick2019}} \href{https://github.com/mlco2/codecarbon}{\ref*{link-lottick2019-bis}} & 44 \\
\bottomrule
\end{longtable}

%% file: tables/YY-analytical-and-data-based-estimation-model.tex
\paragraph{YY studies in the group ``analytical and data-based estimation model''} 
 \label{tab:YY-analytical-and-data-based-estimation-model} 
\begin{longtable}{|>{\raggedright\arraybackslash}p{2cm}|p{5.9cm}|>{\raggedright\arraybackslash}p{1.85cm}|>{\raggedright\arraybackslash}p{1.95cm}|>{\raggedright\arraybackslash}p{1.5cm}|p{0.75cm}|}
\toprule
\bfseries study & \bfseries detail & \bfseries target task & \bfseries constraints & \bfseries available & \bfseries cites \\
\midrule 
\endhead
\cite{getzner2023}, 2023, DLEE & (dl-energy-estimator) Linear and polynomial regression models with \acrshort{nn} architecture and \acrshort{mac} count as inputs (layer-wise consumption), accounting for CPU and RAM & Deep \acrshort{nn} inference & no & code (data collection, training), model \href{https://github.com/JohannesGetzner/dl-energy-estimator}{\ref*{link-getzner2023}} & 1 \\
\cite{lahmer2022}, 2022, \acrshort{nu} & Model with \acrshort{mac} count and platform-specific parameters (parameters obtained empirically from data) as inputs, accounting for CPU and RAM (unclear if GPU power is accounted for) & Deep \acrshort{nn} inference (fully connected, convolutional) & Nvidia Jetson edge computer & no & 7 \\
\bottomrule
\end{longtable}

%% file: tables/YY-analytical-estimation-model-and-on-chip-sensors.tex
\paragraph{YY studies in the group ``analytical estimation model and on-chip sensors''} 
 \label{tab:YY-analytical-estimation-model-and-on-chip-sensors} 
\begin{longtable}{|>{\raggedright\arraybackslash}p{2cm}|p{5.9cm}|>{\raggedright\arraybackslash}p{1.85cm}|>{\raggedright\arraybackslash}p{1.95cm}|>{\raggedright\arraybackslash}p{1.5cm}|p{0.75cm}|}
\toprule
\bfseries study & \bfseries detail & \bfseries target task & \bfseries constraints & \bfseries available & \bfseries cites \\
\midrule 
\endhead
\cite{qiu2023}, 2023, \acrshort{nu} & \acrshort{rapl}, \acrshort{nvml}, and model with download/upload speed, ML model size and router power as inputs (for communications), accounting for CPU, RAM, GPU and Wide Area Networking & Federated Learning training & no & no & 42 \\
\cite{budennyy2022}, 2022, Eco2AI & Model with hardware utilization as inputs for the CPU and RAM, \acrshort{nvml} for the GPU, accounting for process-level, for CPU, RAM and GPUs (of the same type) & any & Nvidia GPU, Python & PyPI package, code \href{https://github.com/sb-ai-lab/Eco2AI }{\ref*{link-budennyy2022}} & 35 \\
\bottomrule
\end{longtable}

%% file: tables/YY-other.tex
\paragraph{YY studies without group} 
 \label{tab:YY-other} 
\begin{longtable}{|>{\raggedright\arraybackslash}p{2cm}|p{5.9cm}|>{\raggedright\arraybackslash}p{1.85cm}|>{\raggedright\arraybackslash}p{1.95cm}|>{\raggedright\arraybackslash}p{1.5cm}|p{0.75cm}|}
\toprule
\bfseries study & \bfseries detail & \bfseries target task & \bfseries constraints & \bfseries available & \bfseries cites \\
\midrule 
\endhead
\cite{ortega2023}, 2023, \acrshort{nu} & Design method of an interface for external power meters, accounting for whole system & any & no & no & 1 \\
\cite{montgomerie-corcoran2021}, 2021, Pommel & Based on Ramulator, Cacti-io and DRAMPower, with memory access traces (from Ramulator) and accelerator specifications as inputs, accounting for off-chip memory on \acrshort{cnn}s accelerator & \acrshort{cnn} inference & \acrshort{cnn} accelerator & code \href{https://github.com/AlexMontgomerie/pommel}{\ref*{link-montgomerie-corcoran2021}} & 0 \\
\cite{wang2021}, 2021, Accelergy & (Gem5-Accelergy-system) Based on MacPAT (simulator) and Timeloop tools, accounting for CPU, RAM, accelerators and data transfer between them & any (tested on Deep \acrshort{nn} inference) & no & code \href{https://github.com/Accelergy-Project/accelergy}{\ref*{link-wang2021}} & 4 \\
\bottomrule
\end{longtable}

%% file: tables/NY-timeline.tex
\begin{tikzpicture} 
\draw[->, thick] (0,0) -- (15,0); 
\foreach \x/\year in {0/2016, 2/2017, 4/2019, 6/2020, 8/2021, 10/2022, 12/2023} { 
    \draw[shift={(\x,0)}, color=black] (0pt,2pt) -- (0pt,-2pt); 
    \node[below] at (\x, 0) {\year}; 
    } 
\node[above right, align=left, rotate=45] at (0,0) {\cite{li2016a}}; 
\node[above right, align=left, rotate=45] at (2,0) {\cite{rungsuptaweekoon2017}}; 
\node[above right, align=left, rotate=45] at (4,0) {\cite{wang2019}\\ \cite{strubell2019}, \cite{mcintosh2019}}; 
\node[above right, align=left, rotate=45] at (6,0) {\cite{rodrigues2020}, \cite{jurj2020}\\ \cite{holly2020}, \cite{yao2021}}; 
\node[above right, align=left, rotate=45] at (8.3,0) {\cite{arnautovic2021}, \cite{guo2021}\\ \cite{naidu2021}, \cite{desislavov2021}\\ \cite{sun2021}, \cite{hesse2021}, \cite{canilang2021}}; 
\node[above right, align=left, rotate=45] at (10,0) {\cite{trihinas2022}\\ \cite{machado2022}, \cite{hampau2022}}; 
\node[above right, align=left, rotate=45] at (12.3,0) {\cite{islam2023}, \cite{caspart2023}\\ \cite{escribano2023}, \cite{hauschild2023}\\ \cite{tekin2023}, \cite{wu2023}}; 
\end{tikzpicture}

%% file: tables/NY-description.tex
\ref{tab:NY-measurement} for the group `measurement', \ref{tab:NY-analytical-estimation-model} for the group `analytical estimation model', \ref{tab:NY-on-chip-sensors} for the group `on-chip sensors', \ref{tab:NY-data-based-estimation-model} for the group `data-based estimation model', \ref{tab:NY-other} for studies without group.%

%% file: tables/NY-measurement.tex
\paragraph{NY studies in the group ``measurement''} 
 \label{tab:NY-measurement} 
\begin{longtable}{|>{\raggedright\arraybackslash}p{0.85cm}|>{\raggedright\arraybackslash}p{4cm}|>{\raggedright\arraybackslash}p{6cm}|>{\raggedright\arraybackslash}p{3.75cm}|}
\toprule
\bfseries study & \bfseries detail & \bfseries ML task & \bfseries setup \\
\midrule 
\endhead
\cite{hauschild2023}, 2023 & \acrshort{epm} JT-TC66C & inference with \acrshort{cnn}s: MobileNetV2, \acrshort{nas}NetMobile, ResNet (50, 101), VGG (16, 19) & edge, server \\
\cite{trihinas2022}, 2022 & \acrshort{epm} & inference with \acrshort{cnn} for object detection (on ImageNet) & edge  \\
\cite{machado2022}, 2022 & \acrshort{epm}/sensors on the board & inference with YOLOv5 for object classification & edge (Nvidia Jetson Nano board) \\
\cite{hampau2022}, 2022 & \acrshort{epm} Monsoon's High Voltage Power Monitor & inference with Image Classifyer (on MNIST, Emotion, CIFAR10) and YOLO & edge \\
\cite{hesse2021}, 2021 & \acrshort{epm} Voltcraft Energy Logger EL4000 (also on-chip sensors if available) & training and inference with \acrshort{cnn}s, including inference with YOLOv3 & FGPA, Apple M1, classical CPU-GPU \\
\cite{wang2019}, 2019 & \acrshort{epm} Monsoon Power Monitor (for mobile) & inference with YOLOv3 & edge server, smartphone \\
\cite{mcintosh2019}, 2019 & \acrshort{epm} INA219 current sensor (within GreenMiner framework) & training and inference with various models (e.g., J48 Decision Tree, ZeroR) & Raspberry Pi for data-collection, Smartphone \\
\cite{rungsuptaweekoon2017}, 2017 & \acrshort{epm} & inference with YOLO & edge (Nvidia Jetson TX1, TX2) \\
\bottomrule
\end{longtable}

%% file: tables/NY-analytical-estimation-model.tex
\paragraph{NY studies in the group ``analytical estimation model''} 
 \label{tab:NY-analytical-estimation-model} 
\begin{longtable}{|>{\raggedright\arraybackslash}p{0.85cm}|>{\raggedright\arraybackslash}p{4cm}|>{\raggedright\arraybackslash}p{6cm}|>{\raggedright\arraybackslash}p{3.75cm}|}
\toprule
\bfseries study & \bfseries detail & \bfseries ML task & \bfseries setup \\
\midrule 
\endhead
\cite{wu2023}, 2023 & power directly approximated by \acrshort{tdp} & training Logistic Regression and \acrshort{cnn} with different algorithms & Intel Xeon Gold 6126 CPU, Nvidia A100 \\
\cite{arnautovic2021}, 2021 & power approximated by calculation of maximum power of the board/chip & inference with MobileNetSSDv2  & edge \\
\cite{desislavov2021}, 2021 & based on \acrshort{flop} count and hardware specifications & inference with Computer Vision and \acrshort{nlp} models & GPUs V100, A100, T4 \\
\cite{canilang2021}, 2021 & Tool of \cite{yang2017} & inference with Deep \acrshort{nn} model for face detection & edge \\
\bottomrule
\end{longtable}

%% file: tables/NY-on-chip-sensors.tex
\paragraph{NY studies in the group ``on-chip sensors''} 
 \label{tab:NY-on-chip-sensors} 
\begin{longtable}{|>{\raggedright\arraybackslash}p{0.85cm}|>{\raggedright\arraybackslash}p{4cm}|>{\raggedright\arraybackslash}p{6cm}|>{\raggedright\arraybackslash}p{3.75cm}|}
\toprule
\bfseries study & \bfseries detail & \bfseries ML task & \bfseries setup \\
\midrule 
\endhead
\cite{caspart2023}, 2023 & internal power sensors of the HoreKa nodes with slurm plugin, \acrshort{nvml} & training and inference with \acrshort{cnn} and Long-Short Term Memory (time series) models & supercomputing system with Intel Xeon Platinum 8368 CPU, Nvidia A100-40 GPUs \\
\cite{escribano2023}, 2023 & Code-Carbon & inference with \acrshort{nlp} model T5-small & application wrapped in Docker and deployed in cloud \\
\cite{tekin2023}, 2023 & Intel-Power-Gadget & traning and inference with ML models (e.g., Logistic Regression, \acrshort{nn}) & edge, IoT, cloud with CPU \\
\cite{naidu2021}, 2021 & Code-Carbon & training (with differential privacy) Bert, Image Classifier, and Reinforcement Learning model for cartpole control & unknown \\
\cite{sun2021}, 2021 & \acrshort{rapl}, \acrshort{nvml} & training LeNet, GoogLeNet, AlexNet, CaffeNet, AlexNet-MNIST & CPU and CPU-GPU platforms (Intel Xeon X5-2650 v3, Nvidia Tesla K80) \\
\cite{jurj2020}, 2020 & powerstat, tegrastats & tranining of VGG-19, InceptionV3, ResNet-50, MobileNetV2 for image classification, inference with MobileNetV2 & edge (Nvidia Jetson TX2), laptop (Nvidia GTX 1060) \\
\cite{holly2020}, 2020 & based on measurements: sensors on the Nvidia Jetson Nano board & inference with MobileNet (V1, V2) and ResNet (18, 50) & Nvidia Jetson Nano \\
\cite{yao2021}, 2020 & \acrshort{nvml} & inference with \acrshort{cnn}s: VGG-16, ResNet-50, Inception-v3 & high-performance GPUs: M40, P4, V100 \\
\cite{strubell2019}, 2019 & \acrshort{rapl}, \acrshort{nvml} & training \acrshort{nlp} models (Transformer T2T, ELMo, BERT, \acrshort{nas}, GPT-2) & accelerators P100, V100, TPUv2, TPUv3 \\
\cite{li2016a}, 2016 & \acrshort{rapl}, \acrshort{nvml} & training and inference with \acrshort{cnn}s: AlexNet v2, OverFeat, VGG-A, and GoogleNet (on ImageNet) & Xeon CPU, K20 GPU, Titan X GPU \\
\bottomrule
\end{longtable}

%% file: tables/NY-data-based-estimation-model.tex
\paragraph{NY studies in the group ``data-based estimation model''} 
 \label{tab:NY-data-based-estimation-model} 
\begin{longtable}{|>{\raggedright\arraybackslash}p{0.85cm}|>{\raggedright\arraybackslash}p{4cm}|>{\raggedright\arraybackslash}p{6cm}|>{\raggedright\arraybackslash}p{3.75cm}|}
\toprule
\bfseries study & \bfseries detail & \bfseries ML task & \bfseries setup \\
\midrule 
\endhead
\cite{rodrigues2020}, 2020 & SyNERGY & inference with \acrshort{cnn}s (e.g., GoogleNet, ResRet50, MobileNet) & Jetson TX1 \\
\bottomrule
\end{longtable}

%% file: tables/NY-other.tex
\paragraph{NY studies without group} 
 \label{tab:NY-other} 
\begin{longtable}{|>{\raggedright\arraybackslash}p{0.85cm}|>{\raggedright\arraybackslash}p{4cm}|>{\raggedright\arraybackslash}p{6cm}|>{\raggedright\arraybackslash}p{3.75cm}|}
\toprule
\bfseries study & \bfseries detail & \bfseries ML task & \bfseries setup \\
\midrule 
\endhead
\cite{islam2023}, 2023 & unknown & inference with frequently used ML models (e.g., Logistic Regression, Multi-Layer Perceptron) & unknown \\
\cite{guo2021}, 2021 & hardware simulation: Design Compiler simulation and CACTI 6.5 & training Binary \acrshort{nn} (on ImageNet): Boolnet, ReActNet, Bi-RealNet, XNOR-Net, BaseNet & 5 accelerators designed in RTL language \\
\bottomrule
\end{longtable}

%% file: tables/YN-timeline.tex
\begin{tikzpicture} 
\draw[dashed, thick] (-1.5,0) -- (1,0); 
\draw[thick] (1,0) -- (13.5,0); 
\draw[dashed, thick] (13.5,0) -- (14,0); 
\foreach \x/\year in {-1/2002, 0.75/2008, 2.5/2009, 4.25/2010, 6.0/2011, 7.75/2012, 9.5/2013, 11.25/2014, 13.0/2015} { 
    \draw[shift={(\x,0)}, color=black] (0pt,2pt) -- (0pt,-2pt); 
    \node[below] at (\x, 0) {\year}; 
    } 
\node[above right, align=left, rotate=45] at (-1,0) {SoftWatt \cite{gurumurthi2002}}; 
\node[above right, align=left, rotate=45] at (0.75,0) {NU \cite{lewis2008}}; 
\node[above right, align=left, rotate=45] at (2.5,0) {NU \cite{singh2009} \cite{ma2009} \cite{spellmann2009}}; 
\node[above right, align=left, rotate=45] at (4.25,0) {LikwidPM \cite{treibig2010}\\ NU \cite{zamani2010}}; 
\node[above right, align=left, rotate=45] at (6.0,0) {NU \cite{suda2011} \cite{basmadjian2011} \cite{piga2011} \cite{matsumoto2011} \cite{chen2011}}; 
\node[above right, align=left, rotate=45] at (7.75,0) {ECAT \cite{chen2012}\\ eprof \cite{schubert2012}}; 
\node[above right, align=left, rotate=45] at (9.5,0) {NU \cite{enam2013} \cite{peng2013} \cite{singh2013}\\ PowerAPI \cite{bourdon2013}, SEFLab \cite{ferreira2013}}; 
\node[above right, align=left, rotate=45] at (11.25,0) {Jalen \cite{noureddine2014a}, BitWatts \cite{colmant2014}\\ NU \cite{storlie2014} \cite{kim2014}, JalenUnit \cite{noureddine2014}}; 
\node[above right, align=left, rotate=45] at (13.0,0) {NU \cite{gutierrez2015a} \cite{foo2015} \cite{harton2015} \cite{gutierrez2015b}\\ E-Surgeon \cite{noureddine2015}};

\draw[dashed, thick] (-1.5,-4.1) -- (0,-4.1); 
\draw[->, thick] (0,-4.1) -- (14,-4.1); 
\foreach \x/\year in {0/2016, 1.75/2017, 3.5/2018, 5.25/2019, 7.0/2020, 8.75/2021, 10.5/2022, 12.25/2023} { 
    \draw[shift={(\x,-4.1)}, color=black] (0pt,2pt) -- (0pt,-2pt); 
    \node[below] at (\x, -4.1) {\year}; 
    } 
\node[above right, align=left, rotate=45] at (0,-4.1) {TEEC \cite{acar2016c} \cite{acar2016b} \cite{acar2016a}\\ NU \cite{li2016b} \cite{hankel2016} \cite{veni2016} \cite{park2016}}; 
\node[above right, align=left, rotate=45] at (1.75,-4.1) {NU \cite{liu2017} \cite{alzamil2017} \cite{jiang2017}\\ Powerstat \cite{becker2017}}; 
\node[above right, align=left, rotate=45] at (3.5,-4.1) {TVAKSHAS \cite{naren2018}\\ NU \cite{fu2018} \cite{dutta2018}}; 
\node[above right, align=left, rotate=45] at (5.25,-4.1) {}; 
\node[above right, align=left, rotate=45] at (7.0,-4.1) {NU \cite{lin2020} \cite{kloh2020} \cite{karantoumanis2020}}; 
\node[above right, align=left, rotate=45] at (8.75,-4.1) {NU \cite{shahid2021b} \cite{shahid2021a} \cite{aboubakar2021}\\ Phantom \cite{montanana-aliaga2021}}; 
\node[above right, align=left, rotate=45] at (11.15,-4.1) {PMT \cite{corda2022}\\ MuMMI \cite{wu2022}, DeepPM \cite{shim2022}\\ NU \cite{morlans2022}, PJ,JJX \cite{noureddine2022}}; 
\node[above right, align=left, rotate=45] at (12.25,-4.1) {NU \cite{alavani2023}, ESAVE \cite{pathania2023}}; 
\end{tikzpicture}

%% file: tables/YN-description.tex
\ref{tab:YN-measurement} for the group `measurement', \ref{tab:YN-analytical-estimation-model} for the group `analytical estimation model', \ref{tab:YN-data-based-estimation-model} for the group `data-based estimation model', \ref{tab:YN-on-chip-sensors} for the group `on-chip sensors', \ref{tab:YN-hybrid} for studies in a mixed group, \ref{tab:YN-other} for studies without group.%

%% file: tables/YN-measurement.tex
\paragraph{YN studies in the group ``measurement''} 
 \label{tab:YN-measurement} 
\begin{longtable}{|>{\raggedright\arraybackslash}p{2.75cm}|p{11.75cm}|p{0.7cm}|}
\toprule
\bfseries study & \bfseries detail & \bfseries cites \\
\midrule 
\endhead
\cite{hankel2016}, 2016, \acrshort{nu} & sensors into motherboard power grid to measure power draw of components, for CPU, RAM, Network, Disk, Power Supply and System & 0 \\
\cite{enam2013}, 2013, \acrshort{nu} & \acrshort{epm} based on the Arduino board, for the whole system & 0 \\
\cite{ferreira2013}, 2013, SEFLab &  (Software Energy Footprint Lab) based on separate measurements of CPU, RAM, Fans, Disk and Motherboard, for execution of a software; \textbf{available:} code \href{https://github.com/SEFLab}{\ref*{link-ferreira2013}} & 64 \\
\cite{piga2011}, 2011, \acrshort{nu} & custom made board measures the computer power via current transducers, a data acquisition device, and a software that controls the framework, for the Disk, CPU and Motherboard & 8 \\
\cite{matsumoto2011}, 2011, \acrshort{nu} & clamp meter, for GPU & 1 \\
\cite{chen2011}, 2011, \acrshort{nu} & an analytical estimation model is proposed, but missing values for the model's weights. Work supported by observations from an \acrshort{epm} (32A PDU gateway from Schleifenbauer), for VMs & 100 \\
\bottomrule
\end{longtable}

%% file: tables/YN-analytical-estimation-model.tex
\paragraph{YN studies in the group ``analytical estimation model''} 
 \label{tab:YN-analytical-estimation-model} 
\begin{longtable}{|>{\raggedright\arraybackslash}p{2.75cm}|p{11.75cm}|p{0.7cm}|}
\toprule
\bfseries study & \bfseries detail & \bfseries cites \\
\midrule 
\endhead
\cite{liu2017}, 2017, \acrshort{nu} & formula based on the amount of consumed energy represented by the static features of source codes and hardware specifications, for a computing task in the cloud, accounts for CPU, RAM and Disk & 13 \\
\cite{acar2016c}, 2016, TEEC &  (Tool to Estimate Energy Consumption) based on information from the Sigar library, for CPU, RAM and Disk & 2 \\
\cite{acar2016b}, 2016, TEEC & model based on hardware utilization, for CPU, RAM, Disk, Network & 12 \\
\cite{park2016}, 2016, \acrshort{nu} & model with utilization rates as inputs, for CPU, RAM, Disk, Mainboard, CPU cooler, Case cooler, and Optical Disc Drive & 3 \\
\cite{acar2016a}, 2016, TEEC & model with hardware utilization as input, for processes, accounts for CPU, RAM and Disk & 29 \\
\cite{noureddine2015}, 2015, E-Surgeon & based on PowerAPI and Jalen, for java classes and methods, accounting for CPU and Network & 80 \\
\cite{noureddine2014a}, 2014, Jalen & based on hardware utilization and power estimation of PowerAPI, for java code on CPU and Network & 16 \\
\cite{peng2013}, 2013, \acrshort{nu} & model based on specific \acrshort{pmc}, for CPU, RAM, Disk, I/O Controller & 11 \\
\cite{bourdon2013}, 2013, PowerAPI & model with hardware utilization, frequency, voltage and specifications as inputs, for processes on CPU, RAM and Disk; \textbf{available:} package, code \href{https://github.com/powerapi-ng/powerapi}{\ref*{link-bourdon2013}} & 94 \\
\cite{basmadjian2011}, 2011, \acrshort{nu} & based on hardware utilization and specifications, for a server, accounting for CPU, RAM, Disk, Mainboard, Network, Fan and Power Supply & 130 \\
\cite{singh2009}, 2009, \acrshort{nu} & model based on \acrshort{pmc} and temperature, for CPU cores & 9 \\
\cite{spellmann2009}, 2009, \acrshort{nu} & based on server power metrics, utilization rates and PUE, for servers & 15 \\
\bottomrule
\end{longtable}

%% file: tables/YN-data-based-estimation-model.tex
\paragraph{YN studies in the group ``data-based estimation model''} 
 \label{tab:YN-data-based-estimation-model} 
\begin{longtable}{|>{\raggedright\arraybackslash}p{2.75cm}|p{11.75cm}|p{0.7cm}|}
\toprule
\bfseries study & \bfseries detail & \bfseries cites \\
\midrule 
\endhead
\cite{alavani2023}, 2023, \acrshort{nu} & ML model (XGBoost) with code features and minimal runtime information as inputs, for CUDA program on GPU & 1 \\
\cite{pathania2023}, 2023, ESAVE &  (Estimating Server And Virtual machine Energy) ML model (XGBoost) with hardware specifications and CPU utlization as inputs, for bare-metal servers & 0 \\
\cite{wu2022}, 2022, MuMMI &  (Multiple Metrics Modeling Infrastructure) tree/rule-based, nonlinear and linear ML models with \acrshort{pmc} as inputs, for CPU and RAM; \textbf{available:} data are available on request from the authors & 2 \\
\cite{shim2022}, 2022, DeepPM &  (Deep Power Meter) ML model (Transformer) with compiled binary as inputs, for CPU & 1 \\
\cite{morlans2022}, 2022, \acrshort{nu} & ML model (fully connected \acrshort{nn}) with hardware specifications as inputs, for laptops & 0 \\
\cite{shahid2021b}, 2021, \acrshort{nu} & linear models with \acrshort{pmc} as inputs, for CPU and RAM or GPU & 8 \\
\cite{shahid2021a}, 2021, \acrshort{nu} & linear models with \acrshort{pmc} as inputs, for CPU and RAM or GPU & 12 \\
\cite{aboubakar2021}, 2021, \acrshort{nu} & Statistical model ARIMA (Autoregressive Integrated Moving Average) based on hardware specifications and utilization, for CPU and RAM & 1 \\
\cite{lin2020}, 2020, \acrshort{nu} & ML model (based on Elman \acrshort{nn} and Long Short Term Memory \acrshort{nn}) with \acrshort{pmc} as inputs, for server & 40 \\
\cite{kloh2020}, 2020, \acrshort{nu} & ML models (Linear Regression, Decision Tree, Support Vector Machine, \acrshort{nn}) with \acrshort{pmc} as inputs, for CPU, memory, cache, Jetson TX2; \textbf{available:} trained models (no documentation) \href{https://github.com/ViniciusPrataKloh/dissertacao-mestrado}{\ref*{link-kloh2020}} & 4 \\
\cite{karantoumanis2020}, 2020, \acrshort{nu} & ML models (Ordinary least squares linear regression, Lasso, Ridge, Epsilon-support vector, Decision tree, Random forest, k-nearest neighbors, Multi-layer Perceptron) with CPU utilization and RAM, Disk and Network information as input, for a scientific application running in a data center; \textbf{available:} some models (in the article) & 1 \\
\cite{fu2018}, 2018, \acrshort{nu} & ML model (Ridge regression) with information from the Perf tool as inputs, for CPU, RAM and Disk & 7 \\
\cite{dutta2018}, 2018, \acrshort{nu} & ML models (ZeroR, Linear Regression, Sequential Minimal Optimization Regression, K-Nearest Neighbor, Reduced Error Pruning Tree, Bagging, Random Forest, \acrshort{nn}) with GPU frequency, memory frequency and hardware resource utilization levels as inputs, for the GPU & 0 \\
\cite{jiang2017}, 2017, \acrshort{nu} & ML model (Elman Neural Network) with information about the temperature, humidity and hardware utilization as inputs, for a server & 0 \\
\cite{li2016b}, 2016, \acrshort{nu} & nonlinear relation between characteristics of a network representing the software and the power used, for CPU, RAM, Network and Disk & 1 \\
\cite{veni2016}, 2016, \acrshort{nu} & ML model (Support Vector Regression) with \acrshort{pmc} as inputs, for CPU, RAM, Cache and Disk of a virtual machine & 10 \\
\cite{gutierrez2015a}, 2015, \acrshort{nu} & ML model (\acrshort{nn}) with \acrshort{pmc} as inputs, for the CPU & 1 \\
\cite{foo2015}, 2015, \acrshort{nu} & ML model (Evolutionary \acrshort{nn}) with inputs such as the number of Map and Reduce, CPU utilization and file size, for jobs in cloud data center & 29 \\
\cite{harton2015}, 2015, \acrshort{nu} & ML model (Random Forest) with CPU utilization as input, for a whole server & 4 \\
\cite{gutierrez2015b}, 2015, \acrshort{nu} & ML model (Feed-forward \acrshort{nn}) with \acrshort{pmc} as inputs, for CPU & 1 \\
\cite{colmant2014}, 2014, BitWatts & BitWatts - model with \acrshort{pmc} as inputs, for processes, accounts for CPU; \textbf{available:} package, code \href{https://github.com/Spirals-Team/bitwatts}{\ref*{link-colmant2014}} & 0 \\
\cite{storlie2014}, 2014, \acrshort{nu} & hierarchical Bayesian modeling with hidden Markov and Dirichlet process models, for an HPC job & 18 \\
\cite{kim2014}, 2014, \acrshort{nu} & model with operating frequency, number of active cores, number of cache accesses, and number of the last level cache misses as inputs, for CPU and RAM of a server & 26 \\
\cite{singh2013}, 2013, \acrshort{nu} & model (Support Vector Machine) with hardware utilization as inputs, for processes on CPU, RAM, I/O and Network & 18 \\
\cite{chen2012}, 2012, ECAT &  (Energy-Consumption-Analysis-Tool) model based on task performance parameters such as CPU utilization, and hardware and software resources allocated, for data-, computation- or communication-intensive task in the cloud & 76 \\
\cite{schubert2012}, 2012, eprof & model with \acrshort{pmc} as inputs, and stack trace used for attribution of used energy to code locations, on CPU, RAM, Disk and Network & 71 \\
\cite{zamani2010}, 2010, \acrshort{nu} & Autoregressive moving average (ARMA) model with past and present \acrshort{pmc} as intputs, for a server with CPUs & 13 \\
\cite{ma2009}, 2009, \acrshort{nu} & Support Vector regression model with workload signals as inputs, for GPU & 152 \\
\bottomrule
\end{longtable}

%% file: tables/YN-on-chip-sensors.tex
\paragraph{YN studies in the group ``on-chip sensors''} 
 \label{tab:YN-on-chip-sensors} 
\begin{longtable}{|>{\raggedright\arraybackslash}p{2.75cm}|p{11.75cm}|p{0.7cm}|}
\toprule
\bfseries study & \bfseries detail & \bfseries cites \\
\midrule 
\endhead
\cite{corda2022}, 2022, PMT &  (Power Measurement Toolkit) based on \acrshort{rapl} or LIKWID (CPU), \acrshort{nvml} (Nvidia GPU) rocm-smi (AMD GPU), for CPU, RAM, GPU, Xilinx FPGAs, and interface to \acrshort{epm}s; \textbf{available:} package, code \href{https://git.astron.nl/RD/pmt}{\ref*{link-corda2022}} & 1 \\
\cite{montanana-aliaga2021}, 2021, Phantom & if not available: analytical estimation model -- based on computation load and hardware specifications, for application or whole system, accounting for CPU, RAM, I/O, GPU, and Network, FGPA, or Embedded Device with a power measurement kit; \textbf{available:} code & 6 \\
\cite{becker2017}, 2017, Powerstat & based on the Power Supply Class of the Linux kernel (exposes information about the power supply to user space), for the whole computer; or \acrshort{rapl}, for the CPU; \textbf{available:} package, code \href{https://github.com/ColinIanKing/powerstat}{\ref*{link-becker2017}} & 5 \\
\cite{treibig2010}, 2010, LikwidPM &  (LIKWID-powermeter) based on \acrshort{rapl}, for CPU and RAM; \textbf{available:} package, code \href{https://github.com/RRZE-HPC/likwid}{\ref*{link-treibig2010}} & 706 \\
\bottomrule
\end{longtable}

%% file: tables/YN-hybrid.tex
\paragraph{YN studies in a mixed group} 
 \label{tab:YN-hybrid} 
\begin{longtable}{|>{\raggedright\arraybackslash}p{2.75cm}|p{11.75cm}|p{0.7cm}|}
\toprule
\bfseries study & \bfseries detail & \bfseries cites \\
\midrule 
\endhead
\cite{noureddine2022}, 2022, PJ,JJX & data-based estimation model and on-chip sensors --  (PowerJoular, JoularJX) PJ: polynomial regression model with CPU cycles as inputs or on-chip sensors, for CPU and GPU on PCs, servers and single-board computer; JJX: based on PJ and a regression model with CPU utilization as inputs, for java methods; \textbf{available:} package, code \href{https://github.com/joular}{\ref*{link-noureddine2022}} & 13 \\
\cite{naren2018}, 2018, TVAKSHAS & analytical estimation model and on-chip sensors: for the difference between actual power draw and utilized power draw -- based on Perf, \acrshort{rapl} and measurement of power draw of CPU, for the CPU & 2 \\
\cite{alzamil2017}, 2017, \acrshort{nu} & measurement and analytical estimation model -- \acrshort{epm} and model with harware utilization as input, for VMs & 3 \\
\cite{suda2011}, 2011, \acrshort{nu} & analytical and data-based estimation model -- based on measured hardware power parameters and on hardware specifications, for SIMD computing task running on CPU and GPU & 0 \\
\cite{lewis2008}, 2008, \acrshort{nu} & analytical and data-based estimation model -- linear regression model with \acrshort{pmc} as inputs, system-wide for servers & 178 \\
\bottomrule
\end{longtable}

%% file: tables/YN-other.tex
\paragraph{YN studies without group} 
 \label{tab:YN-other} 
\begin{longtable}{|>{\raggedright\arraybackslash}p{2.75cm}|p{11.75cm}|p{0.7cm}|}
\toprule
\bfseries study & \bfseries detail & \bfseries cites \\
\midrule 
\endhead
\cite{noureddine2014}, 2014, JalenUnit & data-based estimation model for the energy variation of libraries based on their input parameters -- based on PowerAPI, Jalen, it is an estimation model for the energy variation of libraries based on their input parameters & 37 \\
\cite{gurumurthi2002}, 2002, SoftWatt & analytical estimation model built upon a computer architecture simulator -- system power simulator built on top of SimOS, for application and OS, on CPU, RAM and Caches & 295 \\
\bottomrule
\end{longtable}

%% file: tables/subsidiary_3.tex
\begin{tabular}{|>{\raggedright\arraybackslash}p{5cm}|p{1.6cm}|p{8.55cm}|}
\toprule
general & location & detail \\
\midrule
\cite{marantos2022}, 2022, Energy-Toolbox & \cite{pijnacker2023} & for application or class, on CPU and GPU \\
\cite{marantos2021}, 2021, SKD4ED & \cite{pijnacker2023} & for application \\
\cite{mancebo2018}, 2018, FEETINGS/EET & \cite{pijnacker2023} & for hardware components of a computer \\
\cite{walker2017}, 2017, Powmon & \cite{garcia-martin2019a} & for mobile CPU \\
\cite{rodrigues2017}, 2017, ARM-Streamline & \cite{garcia-martin2019a} & for ARM mobile CPU \\
\cite{nucci2017}, 2017, PETrA & \cite{ergasheva2020} & for method, for mobile applications \\
\cite{pereira2017}, 2017, SPELL & \cite{pijnacker2023} & for method or program on computer \\
\cite{rouhani2016}, 2016, DeLight & \cite{garcia-martin2019a} & for training of feed-forward NN \\
\cite{chowdhury2016}, 2016, GreenOracle & \cite{pijnacker2023} & for application on Android \\
\cite{westfield2016}, 2016, Orka & \cite{pijnacker2023} & for method on Android \\
\cite{couto2015}, 2015, Greendroid & \cite{rieger2017} & for Android programs during unit tests \\
\cite{aggarwal2015}, 2015, Green-Advisor & \cite{ergasheva2020} & for changes of energy profile of an application \\
\cite{cordero2015}, 2015, GreenSoM & \cite{pijnacker2023} & for Java-class \\
\cite{godboley2015}, 2015, Green-JEXT & \cite{pijnacker2023} & for application \\
\cite{wilke2013}, 2013, JouleUnit & \cite{rieger2017} & for unit testing of application on any platform \\
\cite{weaver2012}, 2012, PAPI & \cite{garcia-martin2019a} & based on RAPL, for Intel CPU and RAM \\
\cite{hao2012}, 2012, eCalc & \cite{pijnacker2023} & for method or program on android \\
\cite{ge2012}, 2012, eTune & \cite{pijnacker2023} & for application or class, on CPU in data center/cloud \\
\cite{david2010, rotem2012, hackenberg2015}, 2012, RAPL & \cite{fahad2019} & for Intel CPU and RAM \\
\cite{hindle2012}, 2012, Green-Mining & \cite{rieger2017} & for software with different versions, on whole system \\
\cite{chung2011}, 2011, ANEPROF & \cite{ergasheva2020} & for Java application on Android \\
\cite{zhang2010}, 2010, PowerTutor & \cite{rieger2017} & for android smartphones \\
\cite{do2009}, 2009, pTop & \cite{noureddine2013} & process-level model based on hardware specifications (TDP) and utilization, for CPU, RAM, Network and Disk \\
\cite{wonil2009}, 2009, ePRO-MP & \cite{ergasheva2020} & for multi-threaded application \\
\cite{li2009}, 2009, McPAT & \cite{garcia-martin2019a} & for C application \\
\cite{flinn1999}, 1999, PowerScope & \cite{noureddine2013, pijnacker2023} & process-level or method level \\
\bottomrule
\end{tabular}

%% file: tables/subsidiary_2.tex
\begin{tabular}{|>{\raggedright\arraybackslash}p{3.25cm}|p{1.25cm}|p{6.7cm}|p{1.5cm}|p{1.5cm}|}
\toprule
name & location & detail & code & doc \\
\midrule
Energy-Scopium & \cite{jay2023} & for CPU, RAM and GPU &  &  \href{https://www.denergium.fr/pages/the-energyscopium-software-suite.html}{\ref*{link-d-Energy-Scopium}} \\
Perf & \cite{jay2023} & Performance analysis tool. Notably provides CPU performance counters, and energy consumption (RAPL based) &  \href{https://www.man7.org/linux/man-pages/man1/perf.1.html}{\ref*{link-c-Perf}} &  \\
Scaphandre & \cite{jay2023} & process-level, for Intel CPU, RAM &  \href{https://github.com/hubblo-org/scaphandre}{\ref*{link-c-Scaphandre}} &  \href{https://hubblo-org.github.io/scaphandre-documentation/}{\ref*{link-d-Scaphandre}} \\
Silicon-Labs & \cite{rieger2017} & for methods calls in embedded software &  &  \\
NVML & \cite{fahad2019} & for Nvidia GPU &  &  \href{https://developer.nvidia.com/nvidia-system-management-interface}{\ref*{link-d-NVML}} \\
Intel-SMC & \cite{fahad2019} & (Intel System Management Controller) for Intel Xeon Phi co-processors &  &  \href{https://www.intel.com/content/dam/develop/external/us/en/documents/xeon-phi-coprocessor-system-software-developers-guide.pdf}{\ref*{link-d-Intel-SMC}} \\
Intel-Power-Gadget & \cite{garcia-martin2019a} & based on RAPL, for Intel CPU and RAM &  &  \href{https://www.intel.com/content/www/us/en/developer/articles/tool/power-gadget.html}{\ref*{link-d-Intel-Power-Gadget}} \\
\bottomrule
\end{tabular}

%% file: tables/subsidiary_1.tex
\begin{tabular}{|>{\raggedright\arraybackslash}p{3.2cm}|p{6.5cm}|p{1.5cm}|p{1.5cm}|p{1.5cm}|l}
\toprule
name & detail & code & doc & blog \\
\midrule
Kepler (Kubernetes Efficient Power Level Exporter) & for Kubernetes systems, at process, container, or Kubernetes pod level, on Intel CPU, RAM and Nvidia GPU, or whole system, or whole network of systems &  \href{https://github.com/sustainable-computing-io/kepler}{\ref*{link-c-Kepler}} &  &  \href{https://sustainable-computing.io/}{\ref*{link-b-Kepler}} \\
Tracarbon & for Intel CPU and RAM &  \href{https://github.com/fvaleye/tracarbon}{\ref*{link-c-Tracarbon}} &  \href{https://fvaleye.github.io/tracarbon/documentation/}{\ref*{link-d-Tracarbon}} &  \href{https://medium.com/@florian.valeye/tracarbon-track-your-devices-carbon-footprint-fb051fcc9009}{\ref*{link-b-Tracarbon}} \\
PyJoules & for Intel CPU, RAM, and Nvidia GPU &  \href{https://github.com/powerapi-ng/pyJoules}{\ref*{link-c-PyJoules}} &  \href{https://pyjoules.readthedocs.io/en/latest/}{\ref*{link-d-PyJoules}} &  \\
Powerstat & Intel CPU, RAM, or laptop on battery &  \href{https://github.com/ColinIanKing/powerstat}{\ref*{link-c-Powerstat}} &  \href{https://manpages.ubuntu.com/manpages/bionic/man8/powerstat.8.html}{\ref*{link-d-Powerstat}} &  \\
PowerTOP & notably process- and system-level &  \href{https://github.com/fenrus75/powertop}{\ref*{link-c-PowerTOP}} &  \href{https://manpages.ubuntu.com/manpages/mantic/en/man8/powertop.8.html}{\ref*{link-d-PowerTOP}} &  \\
\bottomrule
\end{tabular}

%% file: tables/linklist.tex
\begin{enumerate}[label={\normalfont \textbf{(L\arabic*)}}] 
 \item \label{link-getzner2023} \url{https://github.com/JohannesGetzner/dl-energy-estimator}
 \item \label{link-you2023} \url{https://github.com/SymbioticLab/Zeus }
 \item \label{link-budennyy2022} \url{https://github.com/sb-ai-lab/Eco2AI }
 \item \label{link-carastan-santos2022} \url{https://github.com/phamthi1812/Benchmark-Tracker}
 \item \label{link-igescu2021} \url{https://github.com/epfl-iglobalhealth/CS433-2021-ecoML}
 \item \label{link-montgomerie-corcoran2021} \url{https://github.com/AlexMontgomerie/pommel}
 \item \label{link-lannelongue2021} \url{https://github.com/GreenAlgorithms/green-algorithms-tool}
 \item \label{link-wang2021} \url{https://github.com/Accelergy-Project/accelergy}
 \item \label{link-trebaol2020} \url{https://github.com/epfl-iglobalhealth/cumulator}
 \item \label{link-anthony2020} \url{https://github.com/lfwa/carbontracker }
 \item \label{link-henderson2020} \url{https://github.com/Breakend/experiment-impact-tracker}
 \item \label{link-lottick2019} \url{https://github.com/responsibleproblemsolving/energy-usage}
 \item \label{link-lacoste2019} \url{https://github.com/mlco2/impact}
 \item \label{link-rodrigues2018} \url{https://github.com/Crefeda/SyNERGY }
 \item \label{link-cai2017} \url{https://github.com/enyac-group/NeuralPower }
 \item \label{link-yang2017} \url{https://energyestimation.mit.edu/}
 \item \label{link-lottick2019-bis} \url{https://github.com/mlco2/codecarbon}
 \item \label{link-corda2022} \url{https://git.astron.nl/RD/pmt}
 \item \label{link-noureddine2022} \url{https://github.com/joular}
 \item \label{link-kloh2020} \url{https://github.com/ViniciusPrataKloh/dissertacao-mestrado}
 \item \label{link-becker2017} \url{https://github.com/ColinIanKing/powerstat}
 \item \label{link-colmant2014} \url{https://github.com/Spirals-Team/bitwatts}
 \item \label{link-bourdon2013} \url{https://github.com/powerapi-ng/powerapi}
 \item \label{link-ferreira2013} \url{https://github.com/SEFLab}
 \item \label{link-treibig2010} \url{https://github.com/RRZE-HPC/likwid}
\end{enumerate}

%% file: tables/subsidiary_linklist_df1.tex
\begin{enumerate}[label={\normalfont \textbf{(L\arabic*)}}] 
\setcounter{enumi}{25} 
 \item \label{link-c-Kepler} \url{https://github.com/sustainable-computing-io/kepler}
 \item \label{link-c-Tracarbon} \url{https://github.com/fvaleye/tracarbon}
 \item \label{link-c-PyJoules} \url{https://github.com/powerapi-ng/pyJoules}
 \item \label{link-c-Powerstat} \url{https://github.com/ColinIanKing/powerstat}
 \item \label{link-c-PowerTOP} \url{https://github.com/fenrus75/powertop}
 \item \label{link-d-Tracarbon} \url{https://fvaleye.github.io/tracarbon/documentation/}
 \item \label{link-d-PyJoules} \url{https://pyjoules.readthedocs.io/en/latest/}
 \item \label{link-d-Powerstat} \url{https://manpages.ubuntu.com/manpages/bionic/man8/powerstat.8.html}
 \item \label{link-d-PowerTOP} \url{https://manpages.ubuntu.com/manpages/mantic/en/man8/powertop.8.html}
 \item \label{link-b-Kepler} \url{https://sustainable-computing.io/}
 \item \label{link-b-Tracarbon} \url{https://medium.com/@florian.valeye/tracarbon-track-your-devices-carbon-footprint-fb051fcc9009}
\end{enumerate}

%% file: tables/subsidiary_linklist_df2.tex
\begin{enumerate}[label={\normalfont \textbf{(L\arabic*)}}] 
\setcounter{enumi}{36} 
 \item \label{link-c-Perf} \url{https://www.man7.org/linux/man-pages/man1/perf.1.html}
 \item \label{link-c-Scaphandre} \url{https://github.com/hubblo-org/scaphandre}
 \item \label{link-d-Energy-Scopium} \url{https://www.denergium.fr/pages/the-energyscopium-software-suite.html}
 \item \label{link-d-Scaphandre} \url{https://hubblo-org.github.io/scaphandre-documentation/}
 \item \label{link-d-NVML} \url{https://developer.nvidia.com/nvidia-system-management-interface}
 \item \label{link-d-Intel-SMC} \url{https://www.intel.com/content/dam/develop/external/us/en/documents/xeon-phi-coprocessor-system-software-developers-guide.pdf}
 \item \label{link-d-Intel-Power-Gadget} \url{https://www.intel.com/content/www/us/en/developer/articles/tool/power-gadget.html}
\end{enumerate}